%% file: main.tex
\documentclass[singlecolumn]{reinpaper}
\input{paper-preamble}
\input{paper-info}
\begin{document}
\maketitle
\begin{abstract}
{\color{black}
Robots deployed in dynamic environments execute changing directed-acyclic-graph (DAG) workloads under limited compute and timing budgets. This article presents \tool, a framework for model scheduling under real-time constraints that combines proportional intermediate deadlines, MIMONet-aware graph refinement, and event-driven synchronization for multi-task inference on resource-constrained robotic platforms. We distinguish three representations: the application DAG, each MIMONet's internal encoder/decoder graph, and the refined scheduling DAG, and make the scope of the timing argument explicit. If the profiled execution bounds, release assumptions, precedence barriers, and per-level capacity premises hold, satisfaction of the assigned sub-deadlines composes to the end-to-end deadline. For non-partitionable tasks, we define a profile-bounded admission rule; before interpreting that rule as empirical timing evidence, we require direct per-task completion checks. \minorrev{We report run-level QoE alongside the offered-node deadline missing rate, making explicit that run completion and successful service of every node are different outcomes.} We propose a comprehensive minibenchmark for evaluating the performance of schedulers under various DNN-driven autonomous driving scenarios. Extensive experiments on Nano Orin, AGX Orin, and M3 Max demonstrate the effectiveness of our proposed method.}
\end{abstract}

\section{Introduction}

{\color{black}
Robotic platforms increasingly execute multiple, often correlated, deep neural networks (DNNs) to deliver perception, prediction, and control. Examples include place and acoustic inference~\citep{bib:MobiSys20:Lee}, multi-camera vehicle inference~\citep{bib:RTSS19:Xiang}, and multimodal affect recognition~\citep{bib:ICMR16:Zhang,bib:TMC21:He}. Multi-DNN pipelines must fit limited memory and compute while satisfying end-to-end timing constraints~\citep{bib:MobiCom18:Fang,bib:MobiSys17:Mathur,kwon2022xrbench,gog2021pylot}. However, even one modern DNN can saturate an embedded accelerator. Thus, co-locating several makes scheduling and precedence management difficult.}

{\color{black}
Weight sharing has emerged as a pragmatic way to lift efficiency for correlated tasks~\citep{bib:NIPS18:He, bib:TMC21:He, bib:MobiSys20:Lee}. Practical pipelines often reuse encoders while branching into task-specific heads: single-input multi-output (SIMONet), multi-input single-output (MISONet), or the combination, multi-input multi-output (MIMONet). MIMONet is particularly attractive on embedded hardware because it reduces parameter count and memory pressure by sharing large backbone features across tasks~\citep{li2023mimonet}. Yet two realities complicate real-time operation on robots. First, \emph{environmental dynamics}, moving obstacles, new objects of interest, or sensor dropouts- change the workload graph online: tasks appear or disappear, dependencies shift, and computational demand fluctuates. Second, heterogeneous sensing and control loops introduce \emph{asynchrony}: even when tasks are correlated, their arrivals and service times rarely align. Generic DAG schedulers, agnostic to MIMONet’s shared-\&-partitionable structure, may leave substantial performance on the table.}

{\color{black}
To address these challenges, we introduce \tool as a model-scheduling framework under real-time constraints for multi-task DNN inference on resource-constrained robots. Its scheduling representation is a refined DAG obtained by expanding a MIMONet stage into shared-encoder invocations and task-specific decoders. Proportional intermediate deadlines and EDF dispatch operate on this refined DAG; event-driven barriers preserve its precedence edges. \minorrev{A task declared non-partitionable remains a single scheduling node with its configured budget; the profile-bounded admission rule states an additional requirement for a deployment claiming a timing guarantee. Under overload, the controller may shed work permitted by its configured protection and urgency rules.} The paper therefore distinguishes conditional analytical composition, profile-bounded empirical timing evidence that requires direct task-level verification, and best-effort overload behavior.

We implement \tool atop PyTorch for single-accelerator systems. The implementation has been exercised on NVIDIA Jetson and Apple M-series platforms and includes a ROS~2 integration. \minorrev{Using the complete median-latency run from each of the five-repetition groups on Nano Orin, AGX Orin, and M3 Max, the extended-minibenchmark comparison gives a mean relative latency reduction of 27.4\% (up to 61.1\%) over EDF across nine platform--scenario configurations. \tool has lower latency in eight configurations and zero Deadline Missing Rate in six selected runs. These are descriptive comparisons of the selected runs, not guarantees across repetitions; Sec.~\ref{sec:extended_overall_effectiveness} reports the aggregation rule, variability, and exceptions.}

\paragraph{Contributions}
\begin{enumerate}[leftmargin=10pt]
\item We provide an explicit mapping from the application DAG through each MIMONet's internal encoder/decoder structure to the refined scheduling DAG, eliminating ambiguity about which graph receives intermediate deadlines and EDF priorities.
\item We state conditional correctness properties for proportional sub-deadline composition and precedence preservation, define a profile-bounded admission rule for non-partitionable tasks under explicit task assumptions, and separate direct empirical timing evidence from aggregate scheduling results and overload shedding.
\item \minorrev{Experiments on Nano Orin, AGX Orin, and M3 Max cover urban navigation, emergency obstacle avoidance, and night driving. A uniform selection from the existing repetitions yields a 27.4\% mean relative latency reduction over EDF and six zero-miss selected configurations, while exposing cases in which refinement, reassignment, or the complete system does not improve every metric. QoE is reported as a run-level lateness indicator alongside the offered-node Deadline Missing Rate.}
\end{enumerate}
}

\section{Background and Motivational Case Study}
\label{sec:motivation}

\begin{figure}[!t]
    \centering
    \includegraphics[width=0.5\textwidth]{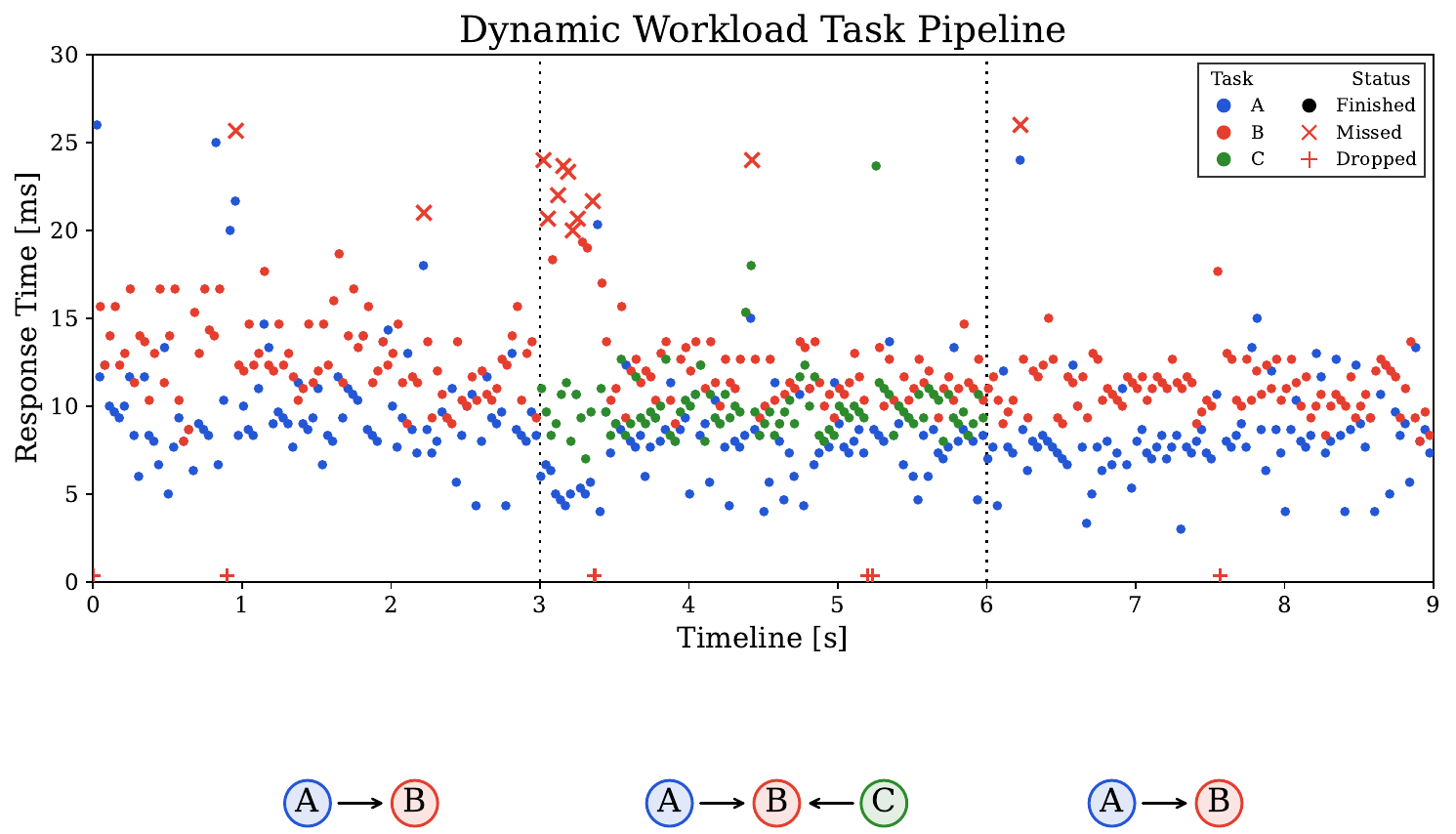}
    \Description{Environmental changes reshape the workload DAG of a running robot.}
    \caption{Environmental changes reshape the workload DAG of a running robot. Color identifies tasks $A$--$C$, while marker shape identifies finished, missed, and dropped outcomes; the red missed and dropped markers use the same colors in the trace and legend. The dotted boundaries mark insertion of task $C$ at 3\,s and its removal at 6\,s.}
    \label{fig:casestudy_1}
    
\end{figure}

\begin{table*}[!htbp]
  \centering
  \caption{Models used in our navigation workloads. Each entry lists the reference implementation and dataset. For embedded realism, some models are downscaled (e.g., YOLOv5n for detection; UNet channels reduced to $1/4$).}
  \label{tab:models}
  \resizebox{0.8\textwidth}{!}{
    \begin{tabular}{c|c|c|c|c|c|c}
    \hline 
    \textbf{Task} & \textbf{Model} & \textbf{Dataset} & \textbf{FLOPs} & \textbf{Params} & \textbf{Mem$_{arm}$} & \textbf{Mem$_{x86}$} \\
    \hline
    Lane Detection & ENet~\citep{paszke2016enet} & KITTI~\citep{geiger2012we} & 4.111G & 3.666M & 3329MB & 3796MB / 2073MB \\
    Segmentation & UNet~\citep{ronneberger2015u}  & KITTI~\citep{geiger2012we}  & 2.655G & 1.943M & 3359MB & 3752MB / 1261MB \\
    Cruise Control & AutoPilot~\citep{bojarski2016end}  & Dave~\citep{pan2017virtual}  & 0.465G  & 2.489M & 3680MB & 3864MB / 1139MB\\
    Object Detection & YOLO~\citep{redmon2016you} & KITTI~\citep{geiger2012we}  & 0.283G & 5.286M & 3593MB & 3754MB / 1159MB\\
    \hline
    Multi-tasks & MIMONet~\citep{li2023mimonet} & KITTI~\citep{geiger2012we} & 6.704G & 10.465M & 3718MB & 3739MB / 1611MB \\
    \hline
    \end{tabular}%
}
  \label{tab:succint}
  
\end{table*}%

\begin{figure*}[!tbp]
\centering
\includegraphics[width=1.0\textwidth]{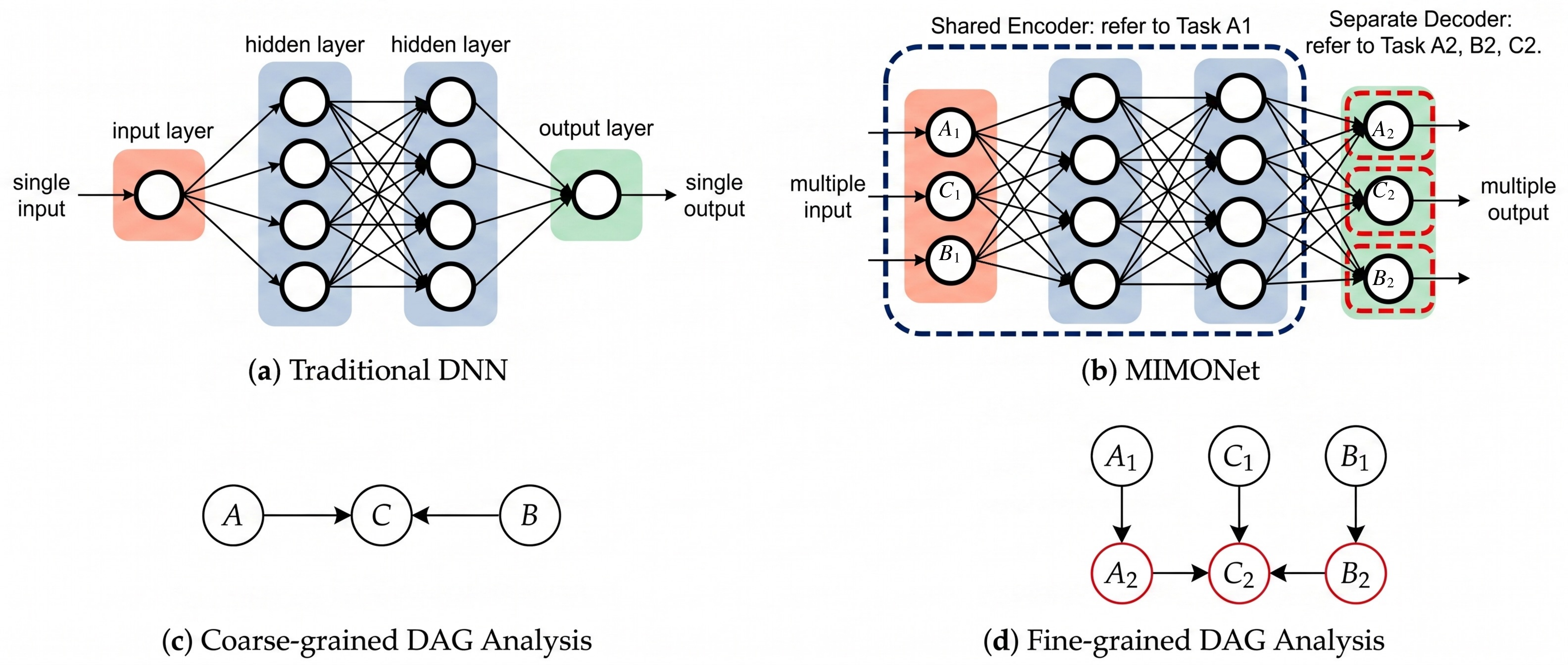}
\caption{\rev{Mapping from model structure to scheduling structure. Panels (a) and (c) show a SISO model and its coarse scheduling representation. In panel (b), plain black identifiers are centered directly in the existing input circles as $A_1$, $C_1$, and $B_1$ from top to bottom and in the existing output circles as $A_2$, $C_2$, and $B_2$. Panel (d) reuses those persistent identifiers for the corresponding refined scheduling nodes; directed edges preserve encoder-before-decoder and application-level precedence.}}
\Description{Four panels compare a traditional single-input model with a multi-input multi-output network and map labeled encoder and decoder invocations to identically labeled scheduling nodes.}

\label{fig:mimonet_arch}
\end{figure*}

{\color{black}
\begin{table}[!t]
\centering
\caption{\textcolor{black}{Mapping from MIMONet model elements in Fig.~\ref{fig:mimonet_arch}(b) to refined scheduling elements in (d) and the constraints each mapping enforces.}}
\label{tab:model_to_sched}
\small
\resizebox{\textwidth}{!}{%
\begin{tabular}{lll}
\toprule
Model element in Fig.~\ref{fig:mimonet_arch}(b) & Scheduling element in (d) & Constraint \\
\midrule
Shared-encoder invocation $A_1/C_1/B_1$ & Same-ID encoder node in (d) & Compatible invocations may merge and execute once per merged group \\
Decoder invocation $A_2/C_2/B_2$ & Same-ID decoder node in (d) & Starts after its encoder event and any application predecessor \\
Tensor flow & Directed scheduling edge & Enforces producer-before-consumer order \\
\bottomrule
\end{tabular}
}
\end{table}
}

This section motivates our design by examining how multi-input multi-output neural networks (MIMONet) behave in robotic settings subject to workload drift and timing pressure. We highlight three stressors—dynamic workload restructuring, structural implications of MIMONet, and asynchronous inference—that together expose limitations of MIMONet-agnostic schedulers.

\subsection{Challenges due to Dynamically Changing Workloads}
\label{sec:case_study_1}

Robots operate amidst moving agents and evolving scenes. These changes can rewire the runtime workload DAG by adding or removing tasks and by altering precedence constraints, which in turn perturb timing behavior.

{\color{black}
Figure~\ref{fig:casestudy_1} reports the recovered historical TurtleBot3/Jetson Nano motivation trace using one documented time unit. The DAG initially contains cruise control $A$ followed by segmentation $B$. Object detection $C$ is inserted at $t=3$\,s, producing $A\!\rightarrow\!B\!\leftarrow\!C$, and is removed at $t=6$\,s. The per-node deadlines for $A$, $B$, and $C$ are 50, 30, and 50\,ms, respectively, under a 130-ms end-to-end budget. Response time is measured from release to completion in milliseconds; a dropped request has no execution latency and is shown on a visible baseline. Counting both late completions and drops as missed outcomes, task $B$ records 5/90 (5.6\%), 12/90 (13.3\%), and 2/90 (2.2\%) missed outcomes in the intervals $[0,3)$, $[3,6)$, and $[6,9)$, respectively. Across all offered tasks in those phases, the corresponding fractions are 5/180 (2.8\%), 12/270 (4.4\%), and 2/180 (1.1\%). The trace therefore illustrates an association between graph mutation and timing pressure. It is a historical motivation example, not admission evidence, and we do not infer a capacity failure from utilization alone.}

\noindent \textbf{Challenge 1.} Dynamic environments induce online changes to task sets and dependencies. Meeting end-to-end deadlines demands a scheduler that adapts to these structural shifts in real time.

\subsection{Challenges due to the MIMONet Architecture}
\label{sec:mimonet_case}

Modern multi-task pipelines commonly share a feature encoder and branch into lightweight, task-specific heads. MIMONet consolidates this pattern into a single model with multiple inputs and outputs (Fig.~\ref{fig:mimonet_arch}), improving memory efficiency and reuse of heavy backbone computation—properties that are highly attractive on embedded GPUs.

These same properties introduce distinct scheduling considerations:

\begin{itemize}[leftmargin=10pt]
\item \textbf{Memory efficiency with shared parameters.} Even minimal MIMONet configurations add only modest parameters relative to a comparable set of independent DNNs, while significantly reducing working-set size on both ARM and x86 platforms (Table~\ref{tab:models}). This makes MIMONet a natural fit for memory-constrained deployments.
\item \textbf{Naturally partitionable computation.} A shared encoder followed by task-specific decoders yields a structure that maps cleanly to finer DAGs: encoder sub-tasks can be shared across multiple heads, whereas decoders remain independent (Fig.~\ref{fig:mimonet_arch}(b,d)). Leveraging this partitionability enables co-scheduling to amortize encoder work, but it also raises new questions about when to merge or split sub-tasks and how to assign intermediate deadlines across the refined graph.
\end{itemize}

\noindent \textbf{Challenge 2.} While MIMONet’s structure saves memory and computation, it also requires schedulers to reason about encoder/decoder sharing and graph granularity. MIMONet-agnostic policies miss opportunities to reduce recomputation or may introduce avoidable stalls.

\begin{figure*}[!tbp]
\centering
\includegraphics[width=\textwidth]{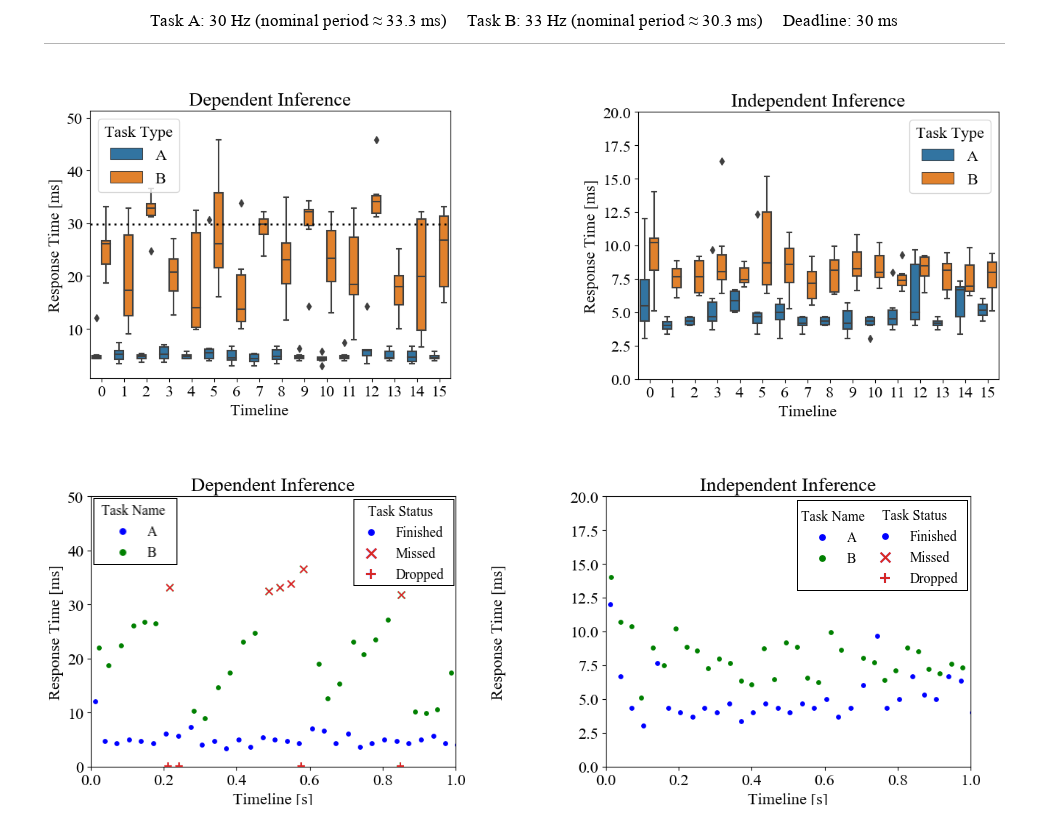}
\Description{Response-time distributions and task outcomes for independent and dependent asynchronous tasks. Task A runs at 30 Hz, task B runs at 33 Hz, and missed and dropped outcomes are marked in red.}
\caption{Asynchrony amplifies timing risk. Task $A$ runs at 30\,Hz (nominal period $\approx 33.3$\,ms), task $B$ runs at 33\,Hz (nominal period $\approx 30.3$\,ms), and both use a fixed 30-ms deadline. We compare response-time distributions and task outcomes under independent and dependent execution. Finished points retain their task colors; missed and dropped outcomes use red markers consistently in the traces and legends.}
\label{fig:casestudy_2}

\end{figure*}

\subsection{Challenges due to Asynchronous Inference}
\label{sec:case_study_2}

Asynchronous inference arises when tasks launch at different rates or with variable service times. This is common in robotics, where perception, planning, and control loops have distinct cadences and where sensors drift relative to one another.

{\color{black}
We evaluate a two-task scenario on TurtleBot3 in a cluttered space. Task $A$ (cruise control) runs at 30\,Hz, whose nominal period is approximately 33.3\,ms; task $B$ (object detection) runs at 33\,Hz, whose nominal period is approximately 30.3\,ms. Both historical traces use a fixed 30-ms per-instance deadline; this deadline is stated separately from the two nominal periods. Figure~\ref{fig:casestudy_2} compares independent execution with the precedence relation $B\!\rightarrow\!A$. In the displayed independent trace, response times remain below 30\,ms and all outcomes are finished. In the dependent trace, several $B$ response times cross 30\,ms and the outcome panel records missed and dropped requests. This descriptive trace motivates dependency-aware scheduling; it is not used to establish the conditional result in Sec.~\ref{sec:schedulability}.}

\noindent \textbf{Challenge 3.} Dependencies within a DAG convert benign rate drift into cascaded blocking. For MIMONet pipelines, encoder sharing further couples tasks, making asynchrony a first-class concern for runtime resource management.

\section{System Model and Problem Formulation}
\label{sec:system_model}

{\color{black}
This section separates the graph supplied by a robotic application from the graph used by the scheduler. It then states what is proved, what must be checked by profiling or admission control, and what remains an empirical overload behavior.

\subsection{Platform and Workload Assumptions}
\label{sec:assumptions}

We target a single accelerator shared by one or more DAG jobs. A scheduling node is non-preemptive once its GPU kernel or runtime call has begun; EDF selects among eligible nodes at dispatch boundaries. CPU preprocessing and data movement may be included in a node's profiled execution bound. The analytical statements below require: (A1) the execution bound assigned to every scheduling node is not exceeded; (A2) releases conform to the admitted arrival model; (A3) all precedence events are enforced; and (A4) every admitted level completes within its allocated interval under the actual co-runner set. A profiled effective-parallelism value $\rho$ may be used as an admission-control estimate, but $\sum c_v/\rho$ alone is not presented as a sufficient EDF schedulability theorem.

\subsection{Three Graph Representations}
\label{sec:task_model}

\noindent\textbf{Definition 1 (Application DAG).}
An application job is
\begin{equation}
G^{A}=(V^{A},E^{A},D),
\label{eq:task_model}
\end{equation}
where $V^{A}$ contains application-visible stages, $E^{A}$ contains their precedence constraints, and $D>0$ is the job's end-to-end relative deadline. A released job has arrival time $a$ and absolute deadline $a+D$.

\noindent\textbf{Definition 2 (MIMONet internal graph).}
For an application stage $x\in V^{A}$ implemented by a MIMONet, $G_x^{M}=(V_x^{M},E_x^{M})$ contains a shared-encoder node $e_x$ and task-specific decoder nodes $q_{x,1},\ldots,q_{x,m}$. Each edge $(e_x,q_{x,j})$ records that decoder $j$ consumes the encoder output. If $x$ is not a MIMONet, its internal graph is the singleton node $x$.

Here ``shared encoder'' names the common encoder implementation and reusable weights, not one permanently collapsed runtime node for every input. Figure~\ref{fig:mimonet_arch} therefore shows the input-specific encoder invocations $A_1$, $C_1$, and $B_1$ as separate scheduler-visible nodes. Compatible invocations may subsequently merge into one execution node; until that decision, their persistent identifiers preserve the correspondence between the model and scheduling graphs.

\noindent\textbf{Definition 3 (Refined scheduling DAG).}
The scheduler operates only on
\begin{equation}
G^{S}=\operatorname{Ref}\!\left(G^{A},\{G_x^{M}:x\in V^{A}\}\right)=(V^{S},E^{S},D).
\label{eq:scheduling_graph}
\end{equation}
The refinement substitutes $G_x^{M}$ for each MIMONet stage $x$, connects every predecessor of $x$ to $e_x$, connects each decoder exit to every successor of $x$, and maps every ordinary or non-partitionable application stage to itself. Thus refinement changes scheduling granularity but preserves every application-level producer--consumer path. From this point onward, $V$ and $E$ abbreviate $V^{S}$ and $E^{S}$.

Each scheduling node is
\begin{equation}
v=(c_v,m_v,\kappa_v,\pi_v,r_v),
\label{eq:node_tuple}
\end{equation}
where $c_v$ is its admitted execution bound, $m_v$ is its accelerator-memory demand, and $r_v$ is its predicted release time for merge eligibility. Its granularity class and role are
\[
\kappa_v\in\{\textsc{Partitionable},\textsc{NonPartitionable}\},\qquad
\pi_v\in\{\textsc{SharedEncoder},\textsc{Decoder},\textsc{Ordinary}\}.
\]
Transfer and event-wait costs are charged to the receiving node's $c_v$.

\begin{table}[!t]
\color{black}
\centering
\caption{\rev{Explicit mapping among the three graph representations.}}
\label{tab:graph_mapping}
\small
\begin{tabular}{p{0.25\columnwidth}p{0.28\columnwidth}p{0.34\columnwidth}}
\toprule
Representation & Node meaning & Used for \\
\midrule
$G^{A}$ & Robot/application stage & External dependencies and end-to-end deadline \\
$G_x^{M}$ & Encoder/decoder component inside stage $x$ & Legal MIMONet decomposition and sharing \\
$G^{S}$ & Dispatchable refined or non-partitionable node & Heights, costs, sub-deadlines, EDF priority, and barriers \\
\bottomrule
\end{tabular}
\end{table}

\subsection{Intermediate Deadlines}
\label{sec:definitions}

\noindent\textbf{Definition 4 (Height and level).}
For the scheduling DAG, a source has height zero and every other node has
\begin{equation}
H(v)=1+\max_{(u,v)\in E}H(u).
\label{eq:height}
\end{equation}
The level set is $\mathcal{L}_h=\{v\in V:H(v)=h\}$, and $H_{\max}=\max_v H(v)$.

\noindent\textbf{Definition 5 (Proportional assignment).}
The interval assigned to scheduling level $h$ is
\begin{equation}
D^{(h)}=D\frac{\sum_{v\in\mathcal{L}_h}c_v}
{\sum_{j=0}^{H_{\max}}\sum_{v\in\mathcal{L}_j}c_v},
\qquad
d_v=a+\sum_{j=0}^{H(v)}D^{(j)}.
\label{eq:proportional}
\end{equation}
By construction, $\sum_hD^{(h)}=D$. This identity allocates the budget; it does not, by itself, prove that the GPU can execute every node by $d_v$.

\noindent\textbf{Definition 6 (Admitted level-service premise).}
\minorrev{Let $F_h$ be the realized time from eligibility of level $h$ until all its nodes complete under the actual dispatcher, co-runners, and resource conflicts. A deployment claiming admission must establish, before dispatch, a service bound $\widehat F_h$ covering that execution envelope and check $\widehat F_h\le D^{(h)}$ for every level, together with A1--A3. The conditional analysis assumes $F_h\le\widehat F_h$; observing $F_h$ after a run is not an admission test. A topology, arrival, or co-runner change requires rechecking the applicable envelope.}

\minorrev{In the evaluated implementation, offline node profiles and frozen workload configurations supply timing estimates, but there is no general online certificate for $\widehat F_h$ and no automatic per-level admission gate. Accordingly, the experiments report empirical scheduling outcomes, including misses, rather than certifying every workload under Definition~6. A deployment requiring the conditional guarantee must additionally provide a platform service analysis or independently validated conservative level-service bounds that include non-preemptive blocking, co-runner interference, and synchronization. An average execution estimate or $\sum c_v/\rho$ is not such a certificate.}

\subsection{Problem Formulation}
\label{sec:problem_formulation}

\noindent\textbf{Refined-DAG Scheduling Problem.}
Given released application DAGs, their internal MIMONet graphs, and a single-accelerator platform, construct $G^{S}$, assign sub-deadlines, and select eligible nodes so that every admitted job completes by $a+D$. If an online mutation or overload invalidates an admission premise, preserve precedence and apply the documented shedding policy; do not interpret that fallback regime as a zero-miss timing guarantee.

\subsection{Conditional Correctness Results}
\label{sec:schedulability}

\noindent\textbf{Proposition 1 (Sub-deadline composition).}
\emph{For a released scheduling DAG $G^{S}$, if every level $h$ becomes eligible no later than $a+\sum_{j<h}D^{(j)}$ and completes within $D^{(h)}$, then every sink completes by $a+D$.}

\noindent\textbf{Proof.}
For $h=0$, all sources are eligible at $a$ and complete by $a+D^{(0)}$. Assume every node below level $h$ completes by $a+\sum_{j<h}D^{(j)}$. Every predecessor of a level-$h$ node lies at a lower level, so the level is eligible by that time. The premise $F_h\le D^{(h)}$ gives completion by $a+\sum_{j\le h}D^{(j)}$. Induction reaches $H_{\max}$; because the allocated intervals sum to $D$, all sinks complete by $a+D$. $\hfill\blacksquare$

\noindent\textbf{Proposition 2 (Refinement preserves application precedence).}
\emph{If Definition~3 is used to construct $G^{S}$ and every edge in $E^{S}$ is enforced, then the projection of any execution trace of $G^{S}$ onto $V^{A}$ respects every edge in $E^{A}$. If the refined graph also satisfies the admitted level-service premise, Proposition~1 applies to the same application deadline $D$.}

\noindent\textbf{Proof.}
An ordinary application node is mapped identically. Replacing a MIMONet stage $x$ creates paths from each predecessor of $x$ through $e_x$ and a decoder exit to every successor of $x$. Therefore every application edge either remains an edge or expands into a directed path. A topological execution of $G^{S}$ consequently projects to a topological execution of $G^{A}$. Refinement does not establish capacity: the second conclusion follows only after the refined costs and conflicts satisfy Definition~6. $\hfill\blacksquare$

\noindent\textbf{Lemma 1 (Event-driven barriers preserve refined edges).}
\emph{Suppose a node is dispatched only after completion events have been observed for all of its immediate predecessors in $E^{S}$. Replacing periodic polling with completion-triggered notification cannot cause a successor to begin before a predecessor completes.}

\noindent\textbf{Proof.}
The readiness predicate is the conjunction of predecessor-completion events. Notification timing changes when the predicate is evaluated, not its truth condition. Hence dispatch implies that every immediate predecessor completed; transitivity gives the full predecessor relation. $\hfill\blacksquare$

\noindent\textbf{Profile-bounded admission rule for non-partitionable tasks.}
A deployment with an independently validated task and contention profile supplies
\begin{equation}
D_{\mathrm{np}}=\mathrm{WCET}_{\mathrm{np}}+\Delta_{\mathrm{cont}}.
\label{eq:nonpart_deadline}
\end{equation}
The rule relies on three assumptions: (NP1) the task follows the profiled code path and remains one scheduling node; (NP2) its execution time in isolation does not exceed $\mathrm{WCET}_{\mathrm{np}}$; and (NP3) the deployed co-runner set, power/clock mode, and contention do not exceed the profiling envelope used to obtain $\Delta_{\mathrm{cont}}$. \minorrev{The margin must cover all additional release-to-completion delay, including queueing and non-preemptive blocking if the deadline is measured from release. A deployment must compare this required budget with the available sub-deadline budget; simply assigning a larger deadline does not make the task feasible. The evaluated code does not separately estimate a certified $\Delta_{\mathrm{cont}}$ or implement this admission check. Equation~\eqref{eq:nonpart_deadline} therefore specifies a conditional deployment requirement, not a verified property of the ratio experiments.} Profiling alone is not a universal analytic guarantee for arbitrary models, kernels, or external co-runners.

\noindent\textbf{Guarantee scope.}
The paper distinguishes three regimes. For an admitted refined DAG satisfying A1--A4 and the level-service premise, Proposition~1 provides conditional end-to-end composition. For a non-partitionable task satisfying NP1--NP3, the profiled $\mathrm{WCET}_{\mathrm{np}}+\Delta_{\mathrm{cont}}$ is an admission threshold; only direct per-task completion checks within that envelope support empirical timing evidence. During transient or sustained overload, the admission premise is violated; the scheduler preserves precedence and priority behavior, but deadline misses may occur.
}

\subsection{Computational Complexity}
\label{sec:complexity}

{\color{black}
Let $|V|$ and $|E|$ denote the total nodes and edges in the current refined scheduling DAGs. Constructing $G^{S}$ and computing heights by topological traversal cost $O(|V|+|E|)$; accumulating level costs and assigning all sub-deadlines cost $O(|V|)$. A direct merge search over an eligible frontier $S$ costs $O(|S|^2)$, or $O(|S|\log|S|)$ when candidates are sorted by predicted release time and a fixed-$\gamma$ window is swept. With predecessor counters, processing a completion costs $O(\operatorname{outdegree}(v))$ and totals $O(|E|)$ over one DAG completion. Scanning a ready queue capped at $Q_{\max}$ costs $O(Q_{\max})$ per overload decision, while updating the $W$-tick persistence counter is $O(1)$. This computation analysis is based on CPU-side orchestration, not actual accelerator schedulability. }

\section{System Design}
\label{sec:design}

\subsection{Overview}
\label{sec:overview}

\begin{figure*}[!t]
    \centering
    \includegraphics[width=\textwidth]{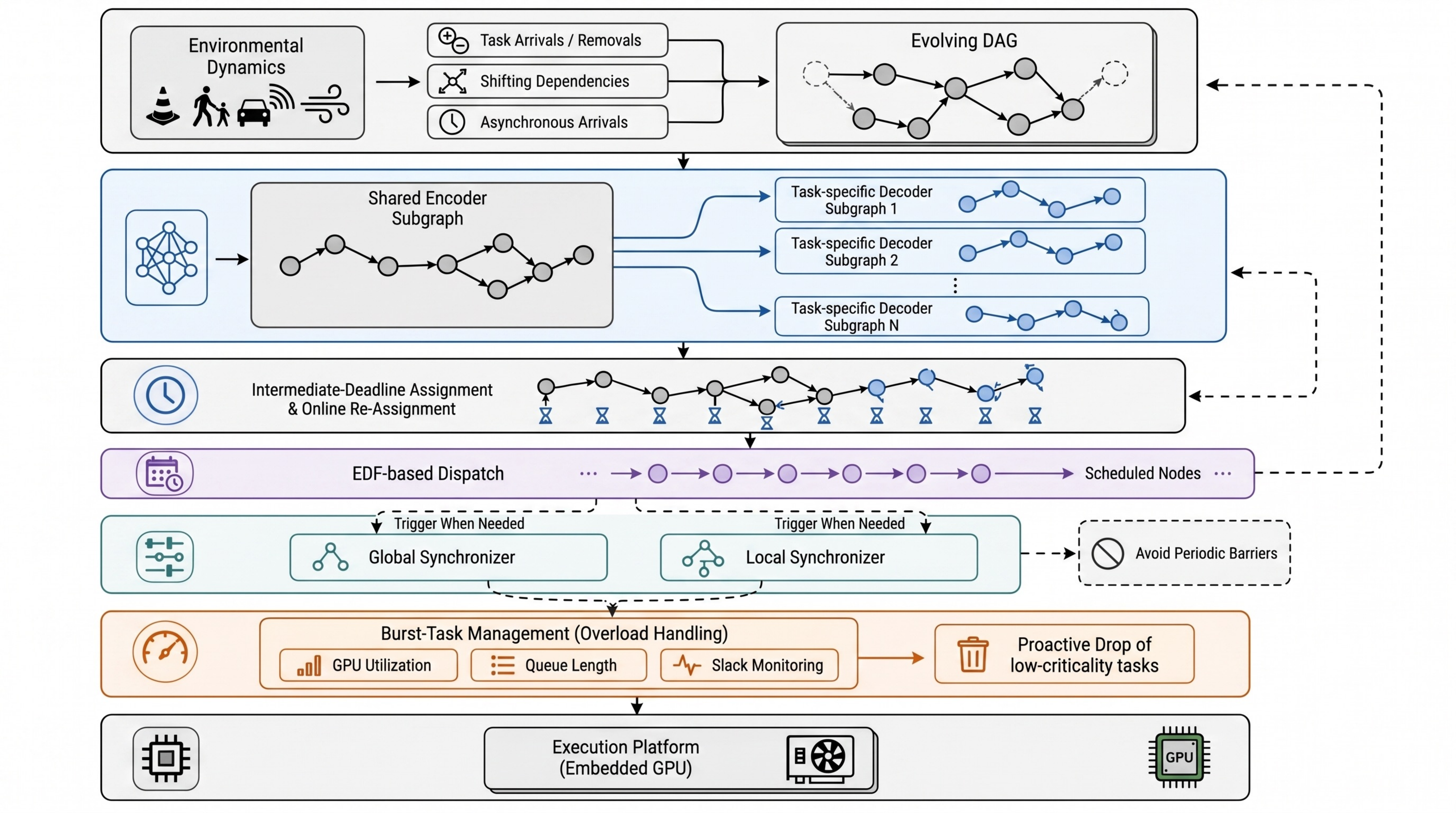}
    \caption{\rev{High-level architecture of \tool. Intermediate-deadline assignment, MIMONet-aware graph refinement/reassignment, and event-driven synchronization operate on the refined scheduling DAG. Timing composition is conditional on A1--A4 and the level-service premise in Definition~6.}}
    \Description{Block diagram of the MERIS runtime, from workload and profile inputs through graph refinement, deadline assignment, dispatch, synchronization, and execution feedback.}
    \label{fig:overview}
    
\end{figure*}

{\color{black}
Motivated by Sec.~\ref{sec:motivation}, \tool targets single-accelerator robotic workloads whose application stages may contain MIMONet encoder/decoder structure. Figure~\ref{fig:overview} follows the three-graph mapping in Table~\ref{tab:graph_mapping}: the runtime first constructs $G^{S}$, then assigns its intermediate deadlines, dispatches eligible nodes, and receives completion events.

The runtime contains five mechanisms. \minorrev{(i) Initial budgets are attached to the materialized scheduling structure, with proportional allocation defined over its levels. (ii) MIMONet refinement exposes shared-encoder dependencies and compatible execution bundling. (iii) Feedback-driven reassignment handles the residual graph at a safe dispatch point; its timing guarantee remains conditional on Definition~6.} (iv) Event-driven synchronization evaluates the same predecessor-completion predicate without periodic polling. (v) Non-partitionable-task handling and overload shedding cover workloads outside the admitted MIMONet regime. Sections~\ref{sec:system_model}--\ref{sec:implementation} define the graph semantics and conditional guarantee scope of these mechanisms.

Meeting all assigned sub-deadlines is sufficient for the end-to-end deadline by Proposition~1. Whether those sub-deadlines are feasible is a separate admission question. The paper therefore does not claim that refinement, EDF, or synchronization alone creates capacity, and it does not claim zero misses during overload.}

\begin{figure*}[!tbp]
    \centering
    \includegraphics[width=\textwidth]{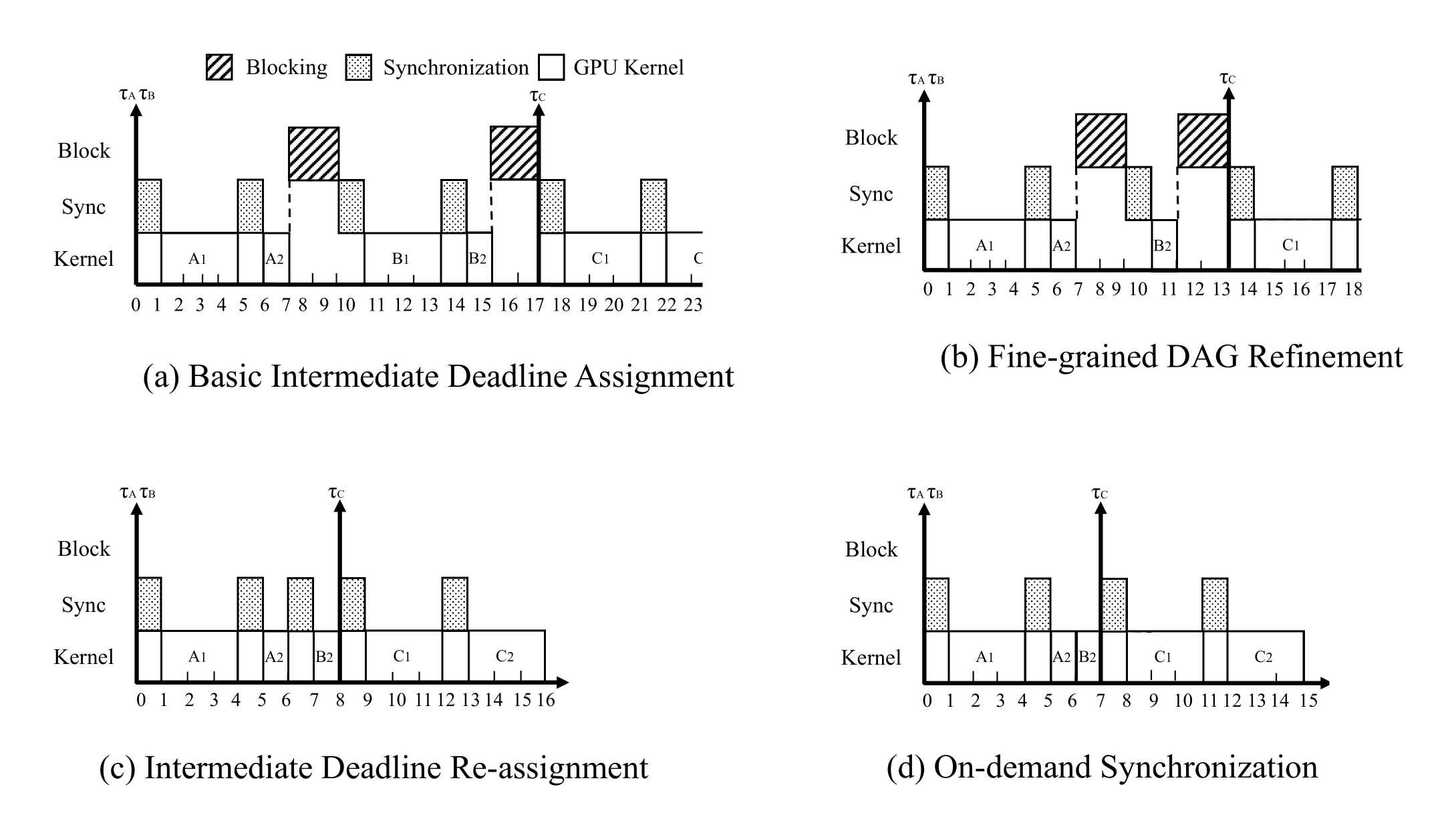}
    \caption{\minorrev{Scheduling operations on a materialized MIMONet structure. Structural expansion in Definition~3 precedes deadline initialization. \textbf{(a)} Assign intervals on the current $G^{S}$. \textbf{(b)} Bundle compatible shared-encoder work while retaining external dependencies and the earliest constituent deadline. \textbf{(c)} Recompute intervals on the residual graph after feedback or a valid graph change; feasibility remains conditional. \textbf{(d)} Use completion events to enforce dependencies without periodic barriers.}}
    \Description{Four panels show proportional deadline assignment, MIMONet graph refinement and compatible merging, deadline reassignment after a graph change, and completion-event synchronization.}
    \label{fig:finer_grained_dag}
    
\end{figure*}

\subsection{Intermediate Deadline Assignment}
\label{sec:intermediate_deadline}

{\color{black}
\minorrev{Definition~5 applies to $G^{S}$, not directly to $G^{A}$ or to an unconnected model-internal diagram. The resulting $D^{(h)}$ values sum to $D$ and produce cumulative absolute sub-deadlines $d_v$. This is a budget decomposition; feasibility is a separate condition in Definition~6. The staged benchmark implementation materializes its task structure and initial node budgets from the TaskConfig before queue construction. Its initial deadline is the configured node release plus the profiled \texttt{WorstRunningTime} scaled by \texttt{DeadlineFactor}; these calibrated initial budgets are not claimed to be a run-time proof of Definition~6. The feedback variants apply the proportional rule to the remaining stage costs and remaining end-to-end budget after completion. Thus graph construction precedes either form of deadline initialization; subsequent compatible merging inherits existing deadlines rather than assigning them on an unexpanded application graph.}

The corrected worked example uses milliseconds. For the chain $A\!\rightarrow\!B\!\rightarrow\!C$ with end-to-end budget 120\,ms, node $A$ is at height 0 with cost 20\,ms, level interval 30\,ms, and cumulative sub-deadline 30\,ms; node $B$ is at height 1 with cost 20\,ms, level interval 30\,ms, and cumulative sub-deadline 60\,ms; and node $C$ is at height 2 with cost 40\,ms, level interval 60\,ms, and cumulative sub-deadline 120\,ms. These intervals sum to the end-to-end budget, and the cumulative values assume release time zero.

The arithmetic illustrates Proposition~1 but does not assert schedulability: admission must still establish that each level completes within its interval under the actual platform and co-runners.}

\subsection{MIMONet-driven DAG Scheduling: Refinement and Re-Assignment}
\label{sec:dynamic_runtime_deadline_assignment}

\paragraph{Fine-grained DAG refinement}
\label{sec:finer_grained_dag_reconstruction}
{\color{black}
MIMONet shares a heavy encoder across task-specific decoders~\citep{caruana1997multitask,maddu2019fisheyemultinet}. Definition~3 expands this structure into $G^{S}$ so that all later heights, merge candidates, costs, and sub-deadlines refer to scheduler-visible nodes. Refinement preserves application precedence by Proposition~2, but it does not prove that the refined work fits the platform; the refined graph must pass the admission premise independently.}

{\color{black}
\minorrev{Structural expansion and execution bundling are distinct. Definition~3 exposes the encoder/decoder dependencies; Algorithm~\ref{alg:algorithm} describes compatible bundling after this structure and its initial deadlines are known. In the benchmark implementation, a stage key identifies the DAG, release instance, and stage, while a handler and its output arguments identify the shared encoder and requested decoders. A bundle is one non-preemptive handler invocation: its encoder runs before the union of requested decoder outputs. Internal encoder--decoder edges are enforced inside that invocation; external dependencies remain dispatch conditions. This specialization does not implement arbitrary kernel-level splitting.}}

\minorrev{Only partitionable invocations from the same stage/instance and compatible handler/input context may be bundled, with predicted release skew $|r_u-r_v|\le\gamma$. Same-stage candidates have no precedence path between them. A bundle retains the union of external dependencies and output identifiers, the constituent identities, and the earliest constituent absolute sub-deadline. Downstream consumers still wait for their required outputs; removing a node during a topological construction is never treated as an execution-completion event. The default $\gamma=100$~ms is configurable and is varied in Sec.~\ref{sec:sensitivity}.}

\begin{algorithm2e}[t!]
\color{revisioncolor}
\DontPrintSemicolon
\SetAlgoLined
\SetKwInOut{Input}{Input}
\SetKwInOut{Output}{Output}
\SetKwInOut{Initialize}{Initialize}
\caption{\minorrev{Dependency-preserving bundling on the refined DAG}}
\label{alg:algorithm}
\Input{Refined scheduling structure $G^{S}$, initial deadlines $d$, skew threshold $\gamma$}
\Output{Execution bundles $B$ and constituent mapping $\phi$}
$B \gets$ singleton invocations of $G^{S}$; $\phi\gets$ identity\;
\ForEach{same-stage, same-instance candidate group in $B$}{
  \ForEach{compatible pair $(u,v)$ with $|r_u-r_v|\le\gamma$}{
    Require partitionability, shared encoder/input context, and no path between $u$ and $v$\;
    $b\gets\textsc{Bundle}(u,v)$; $d_b\gets\min(d_u,d_v)$\;
    Retain the union of external predecessor requirements and output identifiers\;
    Map all constituents of $u,v$ to $b$ in $\phi$; replace $u,v$ by $b$ in $B$\;
  }
}
Dispatch a bundle only after its release and all external dependency events; enforce internal encoder--decoder edges in the handler\;
\Return{$B,\phi$}
\end{algorithm2e}

\paragraph{MIMONet-driven intermediate-deadline reassignment}
{\color{black}
After initial allocation, \tool{} schedules eligible nodes by EDF using their assigned sub-deadlines. At each valid task completion, the execution-cost estimate for that scheduling-node profile is updated from the most recent $k$ observations; the implementation uses a configurable \texttt{MovingAverageWindow}, with default $k=8$, and falls back to the offline-profiled execution bound before observations are available. At the subsequent dispatch-safe reassignment point, completed nodes are removed and Definition~5 is reapplied to the residual $G^{S}$ using the remaining end-to-end budget. The reassigned intervals therefore sum to that residual budget. The moving average changes the online cost estimate, not the end-to-end deadline or the definition of schedulability. Section~\ref{sec:sensitivity} varies $k$ explicitly, and every run records the configured window and the raw reassignment trace. Proposition~1 remains conditional on the residual graph satisfying the admitted level-service premise.}

\subsection{On-demand Synchronization}
\label{sec:implementation}

Synchronization is essential for correctness but expensive if overused. Fixed-interval barriers are simple to implement yet trigger regardless of need, inflating overhead and queuing delay. \tool{} replaces these periodic barriers with \emph{on-demand} events: synchronization occurs when frontier sets complete, when encoder outputs must be shared safely, or when a state transition influences downstream dispatch (Fig.~\ref{fig:finer_grained_dag}{(d)}). Crucially, on-demand synchronization still preserves correctness in the precedence-preservation sense formalised by Lemma~1 in Sec.~\ref{sec:schedulability}: the optimisation only suppresses \emph{unnecessary} barriers (e.g., periodic barriers that fire when no precedence edge is yet ready to cross), never the \emph{required} barriers \minorrev{specified by the conjunction of immediate-predecessor completion events in Lemma~1}, so no node $v_j$ ever begins execution before all its predecessors $\{u : (u, v_j) \in E\}$ have completed.

\begin{figure}[!t]
\centering
\begin{subfigure}[t]{0.62\textwidth}
    \centering
    \includegraphics[width=\textwidth]{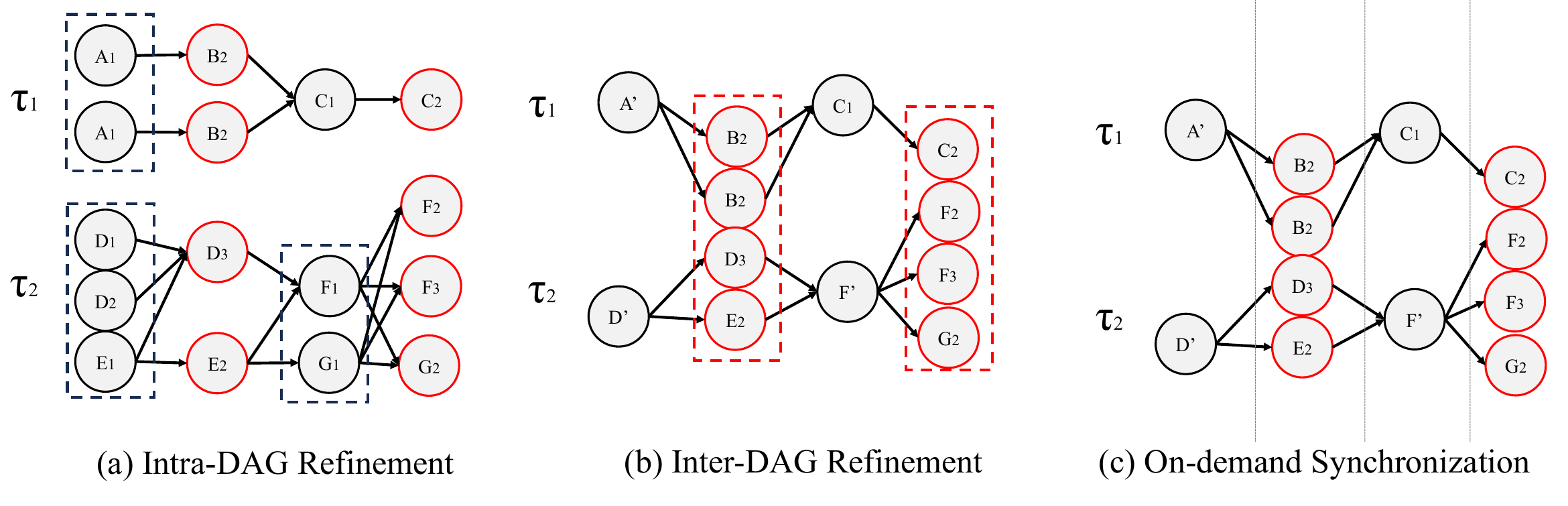}
    \caption{Parallel execution across multiple DAGs. Intra-DAG merges amortize shared encoders; inter-DAG batching of same-height nodes raises parallelism; on-demand sync removes redundant waits.}
    \label{fig:multi_DAG}
\end{subfigure}\hfill
\begin{subfigure}[t]{0.34\textwidth}
    \centering
    \includegraphics[width=\textwidth]{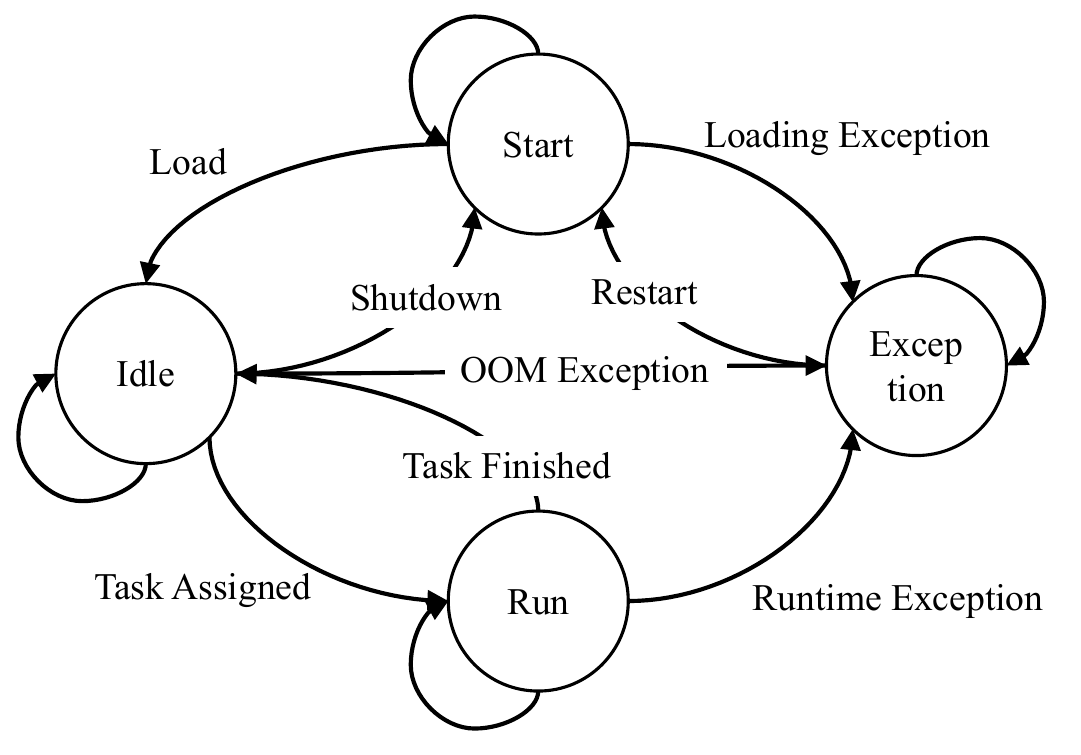}
    \caption{Handler finite-state machine: a state-transition diagram states capture readiness, running, I/O, OOM, and completion to inform scheduling. Transitions are triggered by handler events.}
    \label{fig:fsm}
\end{subfigure}
\caption{Multi-DAG optimizations and control flow in \tool.}
\Description{The left panel shows parallel and merged execution across several DAGs; the right panel shows handler state transitions among readiness, running, input-output, out-of-memory, and completion states.}
\label{fig:dag_and_fsm}

\end{figure}

\noindent\textbf{Global and local synchronizers.}
A global synchronizer coordinates application-wide progress: it polls the scheduler, issues control signals to handlers, and collects completion feedback. Each handler maintains a local synchronizer backed by a compact FSM (Fig.~\ref{fig:fsm}) that tracks internal readiness, memory availability, and I/O states. Local transitions prompt notifications to the controller, enabling timely rescheduling decisions. Both synchronizers use lightweight spin-lock constructs to minimize latency at the expense of modest CPU usage, a trade-off that proved beneficial on embedded SoCs.

\noindent\textbf{Multi-DAG execution.}
In workloads with several DAGs, \tool{} applies intra-DAG refinement to cut redundant compute and then batches independent nodes at the same height across DAGs to exploit parallel resources (Fig.~\ref{fig:multi_DAG}). Synchronization is deferred to height boundaries by default, and only emitted once all nodes at that level complete, further reducing coordination overhead.

\subsection{{Support for Non-partitionable Tasks}}
\label{sec:non_partitionable_tasks}

{\color{black}
\tool{} explicitly supports workloads that contain non-partitionable tasks. The property is part of the workload model: if a task cannot or should not be split by the scheduler, it remains one node in $G^{S}$, participates in the same precedence checks and deadline-ordered queue as other tasks, and \minorrev{uses its configured sub-deadline. The \texttt{IsNonPartitionable} flag excludes it from compatible merging; it does not itself perform the admission test in Eq.~\eqref{eq:nonpart_deadline}.} The evaluation question is how \tool{} schedules such workloads as the proportion of non-partitionable tasks increases; it does not require a separate claim that a particular model or GPU kernel is physically impossible to decompose.

\minorrev{The profiling utility times synchronized encoder-plus-selected-decoder calls in inference mode, after warm-up, and stores the observed maximum multiplied by a fixed margin as \texttt{WorstRunningTime}. For example, the AGX Orin profile uses 20 warm-ups, 200 measured iterations per path, and a 10\% margin (Sec.~\ref{sec:eval_nonpartitionable}). The frozen configuration records the profile identity, platform mode, and timing scale. These shared, fixed inputs make the workload comparison reproducible across methods. The ratio experiment varies which offered nodes are merge-ineligible while keeping their identities and timing inputs matched; its design measures comparative scheduling behavior under those conditions. This finite-sample execution estimate is used to initialize budgets and online cost estimates; the 10\% margin is not a separately measured worst-case contention bound. A certified $\Delta_{\mathrm{cont}}$ would require additional service-envelope evidence covering the deployed co-runners and waiting delays. No such certification is inferred from Table~\ref{tab:nonpartitionable}, which reports comparative scheduling behavior rather than direct verification of Eq.~\eqref{eq:nonpart_deadline}.}}

\subsection{{Burst Task Scenarios Management}}
\label{sec:burst_task_scenarios}

{\color{black}
\textcolor{black}{The overload controller observes GPU utilization $U_{\mathrm{GPU}}$, ready-queue length $Q_{\mathrm{len}}$, and \minorrev{the time remaining to a node deadline.} It invokes \textsc{ProactiveDrop} only after the utilization condition ($U_{\mathrm{GPU}}>\theta_u$) or the queue condition ($Q_{\mathrm{len}}>Q_{\max}$) persists for $W$ consecutive scheduler ticks. The utilization threshold $\theta_u$, the queue threshold $Q_{\max}$, and the persistence window $W$ are platform-specific calibration parameters, not fixed design constants: their concrete per-platform values are disclosed with the burst experiments in Sec.~\ref{sec:eval_burstmanagement}, and their influence is evaluated in Sec.~\ref{sec:sensitivity}. No single configuration of $(\theta_u,Q_{\max},W)$ is treated as universal.}

\minorrev{For the implementation's release-relative deadline-urgency score, let $a_i$ be the configured release timestamp and $b_i=d_i-a_i>0$ its deadline budget; these are distinct from the predicted-release symbol $r_i$ used in merge eligibility. At current timestamp $t$, let $e_i(t)=\max(0,t-a_i)$ and $S_i=b_i-e_i(t)$. The code evaluates:}
\begin{equation}
\minorrev{C_i=\min\!\left(1,\max\!\left(0,1-\frac{S_i}{b_i}\right)\right)
       =\min\!\left(1,\max\!\left(0,\frac{e_i(t)}{b_i}\right)\right).}
\label{eq:crit_score}
\end{equation}
\minorrev{The earlier notation $S_i=d_i-t$ indeed simplifies to $\operatorname{clip}(t/d_i)$ if both quantities are release-relative. Writing $b_i$ and $e_i(t)$ explicitly resolves the ambiguity with the absolute sub-deadline $d_i$ used elsewhere. This is the fraction of the task's own deadline budget that has elapsed, not laxity: it contains neither remaining execution cost nor a class-criticality input. Within the configured droppable set, lower scores identify tasks that have consumed less of their relative deadline budget, so, for tasks with the same protection and threshold settings, threshold-based shedding preferentially retains work with a larger consumed-budget fraction. Normalization allows comparison across different relative deadlines using release/deadline metadata, without an additional remaining-cost prediction in the shedding decision. Saturation prevents distinguishing overdue tasks. This explains the heuristic\textquotesingle s role, but does not establish optimality or a mixed-criticality mode-switch guarantee.}

\minorrev{Protection is configured separately. In the Nano Orin and M3 Max burst paths, a released queued task is shed only when $C_i$ is below its per-task \texttt{DropThreshold}; a zero threshold protects the task because $C_i\ge0$. These paths preserve the order of retained work rather than globally sorting the ready queue by $C_i$. The AGX Orin burst path additionally checks \texttt{Droppable}, protects a merged bundle if any constituent is protected, and limits shedding toward a recovery queue watermark. Its eligible candidates are ordered by burst membership, configured priority, and then increasing $C_i$; during recovery only residual burst work is eligible. Thus the score influences urgency within an explicitly configured shedding policy, while class protection does not follow from Eq.~\eqref{eq:crit_score} alone. Section~\ref{sec:eval_burstmanagement} reports the actual drop composition.}}

\begin{algorithm2e}[!tbp]
\color{revisioncolor}
\DontPrintSemicolon
\SetAlgoLined
\SetKwInOut{Input}{Input}
\SetKwInOut{Output}{Output}

\caption{\minorrev{Integrated task orchestration with explicit graph construction}}
\label{alg:integrated_burst}

\Input{Application DAG $G^A$, internal model graphs, deadline $D$, node profiles/configuration}
\Output{Scheduled Task Execution}

$G^S\gets\operatorname{Ref}(G^A,\{G_x^M\})$ \tcp*{Construct scheduling structure first}
Initialize deadlines on $G^S$ from Definition~5 or the materialized benchmark budgets of Sec.~\ref{sec:intermediate_deadline}\;
$(B,\phi)\gets\textsc{CompatibleBundle}(G^S,d,\gamma)$ \tcp*{Inherit deadlines; Alg.~\ref{alg:algorithm}}
Initialize completion events, handler state, and offered-node outcome records\;
\While{some offered node has no terminal outcome}{
    Process completion, skip, drop, and error events; update constituent records\;
    \If{a valid completion is received and feedback is enabled}{
        Update cost estimates; proportionally reassign the residual stage budgets at a safe dispatch point\;
    }
    Dispatch released dependency-ready bundles by EDF when their handler is available\;
    Check overload on released waiting work and apply the configured protection/threshold rules\;
    Record sheds and expired or dependency-blocked cancellations as non-success outcomes\;
}
\end{algorithm2e}

\section{Evaluation}

\label{sec:evaluation}

{\color{black}
This section evaluates \tool on four NVIDIA embedded platforms through
overall-effectiveness results, a ROS~2 case study, adaptability analysis, and
overhead measurements. Section~\ref{sec:extended_evaluation} reports calibrated
experiments on Nano Orin, AGX Orin, and M3 Max; hardware descriptions and results
are not duplicated across the two sections.}

In this section, we test our full implementation of \tool on top of the PyTorch framework with an extensive set of evaluations. We explore its overall effectiveness under different deadline configurations, from tight to relaxed deadlines, across several hardware platforms with distinct characteristics (Sec.~\ref{sec:overall_effectiveness}). This evaluation aims to demonstrate the general usability and flexibility of \tool in a wide range of scenarios.
Next, we delve into a practical case study, integrating \tool with ROS2 on NVIDIA IoT AI, to understand its effectiveness and adaptability in real-world, complex situations (Sec.~\ref{sec:ros2}). This highlights how seamlessly our solution can be integrated with ROS, an aspect we introduced in the introduction.
Additionally, we conduct a parameter study to demonstrate \tool's flexibility in handling varying deadline settings (Sec.~\ref{sec:adaptability}).
We then evaluate the computational overhead of the \tool framework (Sec.~\ref{sec:overhead}), providing insight into its operational costs. 

\subsection{Experimental Setups}

\label{sec:setup}

This section details a comprehensive evaluation setup, complete with varying hardware platforms, a diverse set of tasks, and a range of deadline configurations, designed to thoroughly test the \tool framework under a variety of conditions. This breadth of testing helps ensure that \tool can deliver consistent, high-quality performance across many potential use-cases in robotic navigation.

\noindent \textbf{Testbeds.} We use pre-trained models or train DNNs following the settings in~\citep{li2023red} on a server with Intel(R) Xeon(R) CPU E5-2650 and a GeForce RTX 2080 Ti GPU. We conduct forward inference experiments on four NVIDIA autonomous embedded platforms as shown in Table~\ref{tab:hardware}.
These platforms are widely used in robotics research~\citep{popov2022nvradarnet} applied as the mainboard of various well-known industrial robots, e.g., Duckiebot~\citep{Duckiebot(DB-J)}, SparkFun Jetbot~\citep{SparkFun_JetBot}, Waveshare Jetbot~\citep{Waveshare_JetBot}, etc. 

\begin{table}[!btp]
  \centering
  \caption{Hardware platforms used in our experiments.}
  \renewcommand\arraystretch{1.3}
  \resizebox{0.7\textwidth}{!}{
    \begin{tabular}{|c|c|c|c|c|}
    \hline
     & {\textbf{Nano}} & {\textbf{TX2}} & {\textbf{Xavier}} & {\textbf{Orin}}\\
    \hline
    {\multirow{3}{*}{CPU}} & 4-core ARM & 6-core ARM & 8-core Armv8.2 & 12-core Armv8.2\\
    & Cortex-A57 CPU & Cortex-A57 CPU & Carmel CPU & Cortex-A78AE \\
    & @ 1.43GHz & @ 1.70GHz & @ 2.03GHz & @ 2.20GHz \\
    \hline
    GPU & NV Maxwell GPU & NV Volta GPU & NV Volta GPU & NV Ampere GPU \\
    \hline
    Memory & 4GB LPDDR4 & 8GB LPDDR4 & 16GB LPDDR4x & 32GB LPDDR5 \\
    \hline
    Storage & 16GB eMMC 5.1 & 32GB eMMC 5.1 & 32GB eMMC 5.1 & 64GB eMMC 5.1 \\
    \hline
    \end{tabular}
  }
  \label{tab:hardware}
  
\end{table}

\noindent \textbf{Metrics.}
\minorrev{We report end-to-end latency, Deadline Missing Rate (DMR), and the legacy Quality of Experience (QoE) score. Their measurement populations differ. End-to-end latency is the elapsed time until the workload run terminates, including accounting for executed, skipped, and dropped work; it is not the time to deliver valid outputs for every offered task. DMR is the fraction of original offered scheduling nodes without a valid on-time output. The reported QoE applies the following continuous lateness function to the workload run, not to an average of individual node outcomes:}

\begin{equation}
QoEScore(r) = \frac{1}{1 + e^{\lambda}[max(0, C_r-S_r)]}
\label{eq:qoe}
\end{equation}

\minorrev{Here $r$ indexes a complete workload run, $C_r$ is its elapsed termination time, and $S_r$ is its configured overall deadline budget, both in seconds from the run start. The code derives $S_r$ as the maximum absolute node deadline generated from the frozen TaskConfig. It does not replace this budget with reassigned deadlines or extend it using injected burst nodes. The metric caller passes one pair $(C_r,S_r)$ to the QoE function; there is no per-node aggregation denominator in the reported score. The configuration supplies $\lambda$ (default 0.05). The equation's functional form is unchanged; the corrected variable definitions describe the values actually passed by the experiment code.}

\minorrev{A late valid completion (\texttt{failure}), a skipped node, a proactive drop, and an execution error each count as one miss in DMR, whose denominator is the original offered-node count before merging. Each constituent of a merged execution retains an outcome record. None of these outcomes is assigned an additional zero or penalty term in run-level QoE: their effect on QoE is only through the run termination time, while their unsuccessful service is captured by DMR. Skipping or dropping work can therefore shorten the run without delivering that work. Equivalently, $\mathrm{DMR}_r=1-N_{\mathrm{on\mbox{-}time},r}/N_{\mathrm{offered},r}$, so unsuccessful work does not disappear from the denominator. The fixed QoE budget supplies the same timing reference to the compared methods under a given configuration; dropping work does not enlarge that budget. QoE equals 1 whenever $C_r\le S_r$, even if individual nodes missed earlier deadlines or were skipped. This is a run-level timing indicator, not an offered-task utility, completion-coverage, safety, or control-quality metric; it must be read together with DMR and, for overload experiments, the drop counts.}

\noindent \textbf{Baselines.}
\minorrev{We compare incrementally enabled configurations of \tool on the offered workload. (1) \textbf{EDF}~\citep{liu2000real} dispatches eligible tasks by their deadlines, without MIMONet-specific bundling or the on-demand synchronization layer. It is a deadline-aware reference policy, not a theoretical lower bound on performance. (2) \textbf{\tool-FG} adds MIMONet-aware refinement and compatible bundling (Definition~3) with EDF dispatch and fixed initial deadlines; the configured benchmark initialization is specified in Sec.~\ref{sec:extended_setup}. Its comparison with EDF includes the effects of both the scheduling representation and shared computation. (3) \textbf{\tool-IDA} adds online residual-budget reassignment (Sec.~\ref{sec:dynamic_runtime_deadline_assignment}), providing an incremental configuration comparison. (4) \textbf{\tool} adds the on-demand synchronization layer (Sec.~\ref{sec:implementation}) to form the complete system. The extended evaluation also includes LLF-DAG and the overload controller where enabled. These configurations expose the effects of enabling successive mechanisms; they do not imply that each addition improves every metric or isolate dispatch policy while holding executed computation constant.} We note that direct comparison against full real-time-DNN frameworks such as S$^3$DNN~\citep{zhou2018s} and ApNet (RTSS'18) is not applicable: those systems target static, single-DNN inference pipelines and do not support MIMONet-style multi-input multi-output structures or graph mutation, so adapting them as baselines would require re-architecting their core dispatch logic and would no longer reflect their published behavior.

\noindent \textbf{Minibenchmark.}
To examine the temporal correctness and throughput performance of the scheduling approach in a dynamic robotic environment, we establish a mini-benchmark employing deep learning tasks as in Tab.~\ref{tab:models}. We configure a sequential three-stage task pipeline for two different driving scenarios: obstacle-free cruising and obstacle cruising. In the obstacle-free case, the three stages consist of (1) lane detection (L), (2) segmentation (S), and (3) cruise control (C). For obstacle cruising, the second stage is modified to include an additional object detection task, becoming segmentation + object detection (S+O), while stages one and three remain the same (L and C, respectively). Each scenario is executed ten times, forming an end-to-end pipeline where each stage must complete before the next begins. The end-to-end deadline for each deep learning task instance is platform-specific and based on worst-case execution time (WCET), tested under two configurations: tight and loose. Specifically, the tight and loose deadlines are set to 9815ms and 11325ms for Nano; 8400ms and 10080ms for TX2; 6500ms and 7315ms for Xavier; and 2400ms and 3600ms for Orin. We set the hyperparameter $\gamma$ as 100ms. Notably, during these mini-benchmark workloads, GPU utilization consistently reaches nearly 100\%, indicating that modern DNN inference is highly GPU-intensive and can saturate the parallel compute capacity available on embedded hardware platforms. {\color{black}The tight and overload configurations are intentional high-workload stress cases: offered demand approaches or exceeds device capacity, so a high Deadline Missing Rate is expected. Their purpose is to expose scheduler behavior under pressure and compare the frequency and severity of misses, not to claim that any scheduler can guarantee all deadlines once demand exceeds capacity.}
 
\begin{figure*}[!t]
    \centering
    \includegraphics[width=\textwidth]{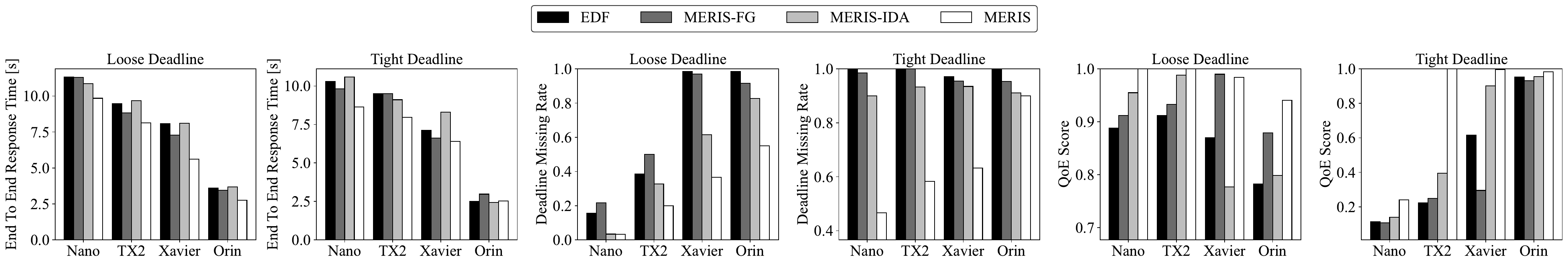}
    \caption{Historical overall-effectiveness results on four resource-constrained robotic platforms. The panels report end-to-end latency, Deadline Missing Rate as a fraction, and QoE under loose and tight deadlines. Platform- and deadline-specific exceptions are discussed in the text; the figure is not evidence of universal dominance or a deadline guarantee.}
    \Description{Six grouped bar charts compare EDF, two MERIS ablations, and MERIS on four platforms for loose and tight end-to-end latency, Deadline Missing Rate, and QoE.}
    \label{fig:overall_effectiveness}
    
\end{figure*}

\begin{figure*}[!t]
    \centering
    \begin{minipage}[b]{0.48\textwidth}
        \centering
        \includegraphics[width=\textwidth]{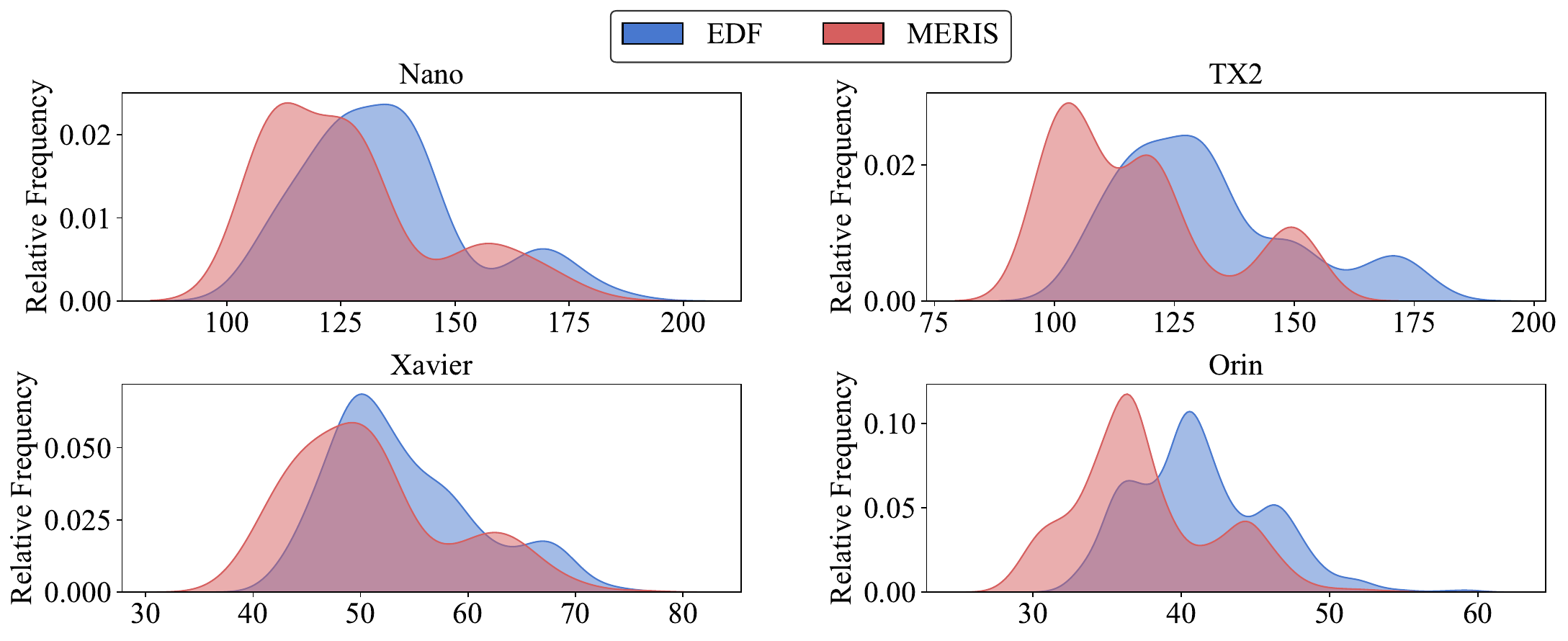}
        \caption{Response time histogram of \tool across different embedded platforms compared to EDF.}
        \label{fig:hist}
    \end{minipage}\hfill
    \begin{minipage}[b]{0.48\textwidth}
        \centering
        \includegraphics[width=\textwidth]{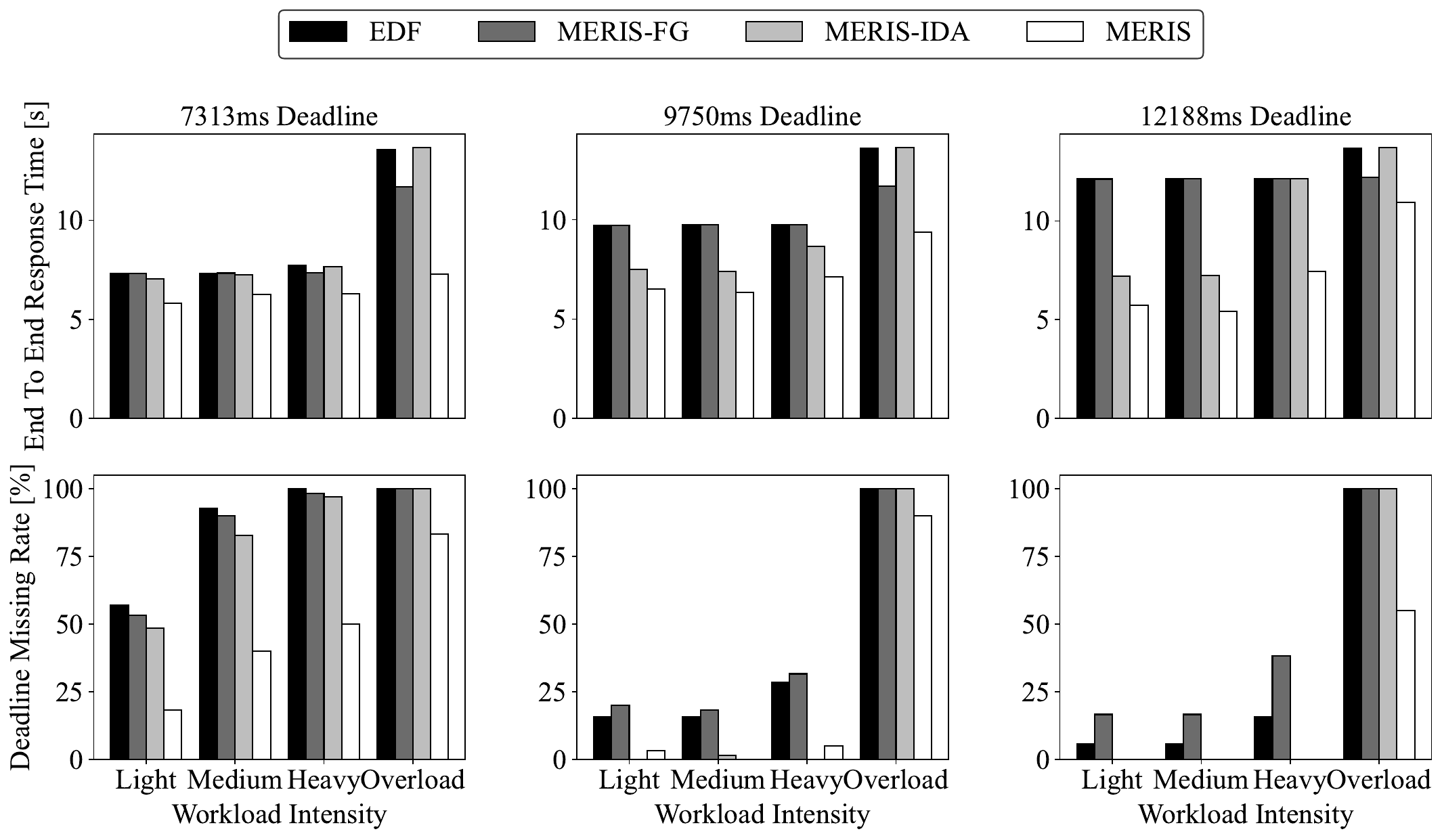}
        \caption{A practical case study on ROS2 involving a camera input with multiple inference outputs on Xavier.}
        \label{fig:ros2_example}
    \end{minipage}
    \Description{The left panel compares response-time distributions for EDF and MERIS on four platforms. The right panel compares latency and Deadline Missing Rate across workload intensities and deadlines in a ROS 2 case study.}
    
\end{figure*}

\subsection{Overall Effectiveness}
\label{sec:overall_effectiveness}

In this section, we report the inherited RTSS evaluation of \tool and its internal components. These historical results are retained for continuity; the calibrated, paired revision campaigns and stronger LLF-DAG comparison appear in Sec.~\ref{sec:extended_evaluation}.

\noindent \textbf{End-to-end Latency.}
As depicted in Fig.~\ref{fig:overall_effectiveness}, \tool has the lowest end-to-end latency in seven of the eight platform--deadline configurations. The exception is the tight-deadline Orin configuration, where \tool-IDA records 2.424~s, EDF records 2.503~s, and \tool records 2.535~s. Averaged over the eight displayed configurations, the reported latency reductions relative to EDF, \tool-IDA, and \tool-FG are 24.7\%, 17.3\%, and 14.2\%, respectively. The task-level histogram in Fig.~\ref{fig:hist} is descriptive: its binned samples show a leftward shift and a lighter observed tail for \tool relative to EDF in the plotted runs. A histogram does not identify exact p99 or p99.9 values and does not establish a worst-case or hard-deadline bound; those conclusions require the admitted premises in Sec.~\ref{sec:schedulability} or direct quantile data.

\noindent \textbf{Real-time Performance.}
Fig.~\ref{fig:overall_effectiveness} reports Deadline Missing Rate as a fraction in the range 0--1 for four embedded devices. Across the eight historical configurations, \tool's average fraction is lower than those of EDF, \tool-IDA, and \tool-FG by the reported 40.5\%, 37.8\%, and 37.5\%, respectively. These are aggregate empirical comparisons, not a zero-miss guarantee: several tight configurations remain far from zero. The ablations suggest that finer-grained partitioning and reassignment alone do not account for the complete-system result, but the figure does not isolate synchronization overhead as a causal mechanism.

\noindent \textbf{Quality of Service.}
Fig.~\ref{fig:overall_effectiveness} also reports QoE as the \minorrev{run-level} lateness indicator defined in Eq.~\eqref{eq:qoe}. \tool has the highest or tied-highest displayed QoE in seven of eight configurations; in the loose-deadline Xavier configuration, \tool-FG records 0.990 and \tool records 0.984. In the TX2 tight configuration, EDF records 0.222 and \tool records 1.000. \minorrev{This difference concerns run-level termination lateness only.} It is not evidence of measured frame pacing, visual stability, control response, trajectory quality, or safety, and a QoE of 1.000 does not imply a Deadline Missing Rate of zero. Accordingly, QoE is interpreted jointly with latency and Deadline Missing Rate throughout the revised evaluation.

\takeaway{Historical on-device evaluation}{The inherited measurements show that the complete system often reduces latency and lateness relative to EDF and its internal ablations, with explicit exceptions above. They motivate the calibrated revision campaigns in Sec.~\ref{sec:extended_evaluation}; they do not establish universal dominance, control stability, or safety.}

\subsection{A Practical Case Study on ROS2}
\label{sec:ros2}

{\color{black}This ROS~2 case study evaluates \tool under varying GPU-interference
intensities using the workload and measurements in Fig.~\ref{fig:ros2_example}.}

We implemented a practical case study on the Robot Operating System 2 (ROS2) utilizing the open-source NVIDIA IoT AI framework~\citep{nvidiaiot}. Specifically, we focused on a realistic robotic application involving a camera input with multiple inference outputs, including lane detection, segmentation, cruise control, and object detection. And we comprehensively evaluate the effect of \tool working under different GPU interference intensities.

To support MIMONet's inference, we modified the open-source library provided by NVIDIA. As shown in Fig.~\ref{fig:ros2_example}, unlike traditional single-task DNNs requiring deployment across multiple ROS nodes, we deployed the MIMONet model on a single ROS node. This modification significantly reduced the communication overhead intrinsic to ROS while enhancing the controllability of each module.
Additionally, we simulate a camera mounted on a car, abstract as a node in ROS2. The MIMONet model processed this data to produce outputs for several deep learning models, as mentioned before. To conduct a fair comparison, we fully implement EDF, \tool{}, and its ablations built on ROS2, strictly following our design.

Figure~\ref{fig:ros2_example} reports the measured ROS~2 case study. Across the displayed interference settings, \tool has a 67.3\% lower Deadline Missing Rate on average and a 32.7\% lower response time than the ROS~2 EDF implementation. These measurements demonstrate that the single-node MIMONet deployment and scheduler execute in the tested ROS~2 pipeline. They are platform- and workload-specific timing results, not evidence of universal dominance, control quality, or safety.

\takeaway{Practical Usability}{This case study demonstrates deployment of \Approach{} in the tested ROS~2 pipeline and reports lower response time and Deadline Missing Rate than its EDF implementation under the displayed interference settings.}

\subsection{Adaptability under Different Scenarios}
\label{sec:adaptability}

\noindent \textbf{Adaptability to variant deadlines.} To investigate the adaptability of the \tool system to various end-to-end deadlines, we conducted a parameter study on the NVIDIA Jetson AGX Xavier platform. This study evaluates the system's performance under different deadlines, ranging from tight to loose. In real-world scenarios, end-to-end deadlines can change due to environmental dynamics, and a robust system should adapt to these changes to meet the required deadlines.
As depicted in Tab.~\ref{tab:variant_deadline}, \tool has the lowest latency in each of the six tested deadline settings; its average latency is 23.2\%, 20.1\%, and 17.1\% lower than EDF, \tool-IDA, and \tool-FG, respectively. This finite Xavier sweep shows that the observed latency benefit persists across the tested deadline range. It does not establish performance for untested workloads or platforms.

\noindent \textbf{Adaptability to variant QoE requirements.} To assess the versatility of our tool in the context of diverse QoE requirements, we maintained the workload intensity at a constant level while adjusting the hyperparameter $\lambda$ of the QoE metrics, as defined in Eq.~\ref{eq:qoe}. This experiment was performed across four unique embedded devices using a stringent tight deadline benchmark as defined in Sec.~\ref{sec:setup}.
Table~\ref{tab:variant_lambda} reports QoE for $\lambda$ values from 0.001 to 10 under a constant workload. \tool has the highest displayed score in the tested cells. Because this is a sweep of one lateness-utility parameter, it shows sensitivity of that metric only; it is not direct evidence of deadline compliance, control behavior, or safety.

\begin{table}[!tbp]
\centering
\caption{End-to-end latency of different approaches under
variant deadline settings on Xavier. A larger deadline-setting ID indicates a looser deadline. The best values are in bold.}
\resizebox{0.6\textwidth}{!}{
\begin{tabular}{c|cccccc}
\hline
Deadline Setting ID & 1 & 2 & 3 & 4 & 5 & 6 \\
\hline
EDF & 5869 & 6807 & 6865 & 7760 & 8737 & 9694 \\
\tool-FG & 5883 & 6006 & 6803 & 7856 & 8723 & 9688 \\
\tool-IDA & 4950 & 4929 & 6691 & 7465 & 8313 & 9045 \\
\tool & \textbf{4455} & \textbf{4435} & \textbf{4456} & \textbf{4500} & \textbf{5930} & \textbf{6779} \\
\hline
\end{tabular}}
\label{tab:variant_deadline}

\end{table}

\begin{table*}[!t]
\centering
\caption{QoE scores of different approaches under variant hyperparameter settings across four embedded platforms. Larger $\lambda$ indicates higher QoE requirements. The best values are in bold.}
\resizebox{0.8\textwidth}{!}{
\begin{tabular}{c|ccccc|ccccc}
\hline
Platform & \multicolumn{5}{c|}{Nano} & \multicolumn{5}{c}{TX2} \\
\hline
$\lambda$ & 0.001 & 0.01 & 0.1 & 1.0 & 10.0 & 0.001 & 0.01 & 0.1 & 1.0 & 10.0 \\
\hline
EDF      & 0.256 & 0.254 & 0.238 & 0.113 & 0.001 & 0.420 & 0.418 & 0.399 & 0.222 & 0.003 \\
\tool-FG   & 0.248 & 0.246 & 0.230 & 0.108 & 0.001 & 0.473 & 0.470 & 0.448 & 0.248 & 0.001 \\
\tool-IDA  & 0.306 & 0.304 & 0.285 & 0.140 & 0.002 & 0.638 & 0.636 & 0.615 & 0.395 & 0.001 \\
\tool      & \textbf{0.461} & \textbf{0.460} & \textbf{0.437} & \textbf{0.240} & \textbf{0.004}
         & \textbf{1.000} & \textbf{1.000} & \textbf{1.000} & \textbf{1.000} & \textbf{1.000} \\
\hline
\hline
Platform & \multicolumn{5}{c|}{Xavier} & \multicolumn{5}{c}{Orin} \\
\hline
$\lambda$ & 0.001 & 0.01 & 0.1 & 1.0 & 10.0 & 0.001 & 0.01 & 0.1 & 1.0 & 10.0 \\
\hline
EDF      & 0.748 & 0.746 & 0.735 & 0.616 & 0.003 & 0.973 & 0.972 & 0.971 & 0.954 & 0.878 \\
\tool-FG   & 0.525 & 0.523 & 0.501 & 0.295 & 0.005 & 0.953 & 0.952 & 0.951 & 0.931 & 0.893 \\
\tool-IDA  & 0.946 & 0.945 & 0.942 & 0.902 & 0.500 & 0.973 & 0.973 & 0.971 & 0.956 & 0.900 \\
\tool      & \textbf{0.999} & \textbf{0.999} & \textbf{0.998} & \textbf{0.996} & \textbf{0.900}
         & \textbf{0.990} & \textbf{0.990} & \textbf{0.989} & \textbf{0.984} & \textbf{0.960} \\
\hline
\end{tabular}}
\label{tab:variant_lambda}

\end{table*}

\takeaway{Tested Adaptability}{The historical deadline sweep reports the lowest latency for \tool at each of six Xavier settings, and the $\lambda$ sweep reports the highest displayed QoE in the tested cells. These bounded sweeps do not imply universal performance or a timing guarantee.}

\subsection{Overhead Analysis}
\label{sec:overhead}

\noindent \textbf{Memory Overhead.}
Table~\ref{tab:combined_overhead}(a) and (b) reports the measured memory overhead of the scheduler and synchronization components, with the scheduler accounting for the smaller share.
However, it is noteworthy that the on-demand synchronization components slightly compromise memory efficiency to enhance latency through reduced context switching between processes. As discussed in Sec.~\ref{sec:implementation}, on-demand synchronization is implemented to decrease code design space coupling, thereby increasing reusability. This approach leads to increased memory overhead due to complex inter-process communication. Future work may consider a potentially more memory-efficient alternative, such as implementing our solution in a multithreading manner, though this may trade off implementation complexity and reusability.
The measured overhead ratios remain low across the four reported hardware platforms and configurations.

\begin{table}[!b]
\centering
\caption{Memory and runtime overhead of applying \Approach{} across platforms. Left: memory overhead (raw and ratio). Right: runtime and profiling overhead.}
\resizebox{\textwidth}{!}{
\begin{tabular}{c|cc|cccc||ccccc}
\hline
& \multicolumn{2}{c|}{(a) Memory Overhead (Raw)} & \multicolumn{4}{c||}{(b) Memory Overhead Ratio} & \multicolumn{5}{c}{(c) Runtime and Profiling Overhead} \\
\hline
& Scheduler & Sync & Nano & TX2 & Xavier & Orin & Platform & Nano & TX2 & Xavier & Orin \\
\hline
Tight & 5 KB & 40.1 MB & 1.00\% & 0.50\% & 0.25\% & 0.13\% & Runtime (ms) & 27.3 & 26.9 & 17.4 & 1.2 \\
Loose & 3 KB & 36.9 MB & 0.92\% & 0.46\% & 0.23\% & 0.12\% & Profiling (s) & 51.1 & 34.3 & 19.1 & 16.1 \\
\hline
\end{tabular}
}
\label{tab:combined_overhead}
\end{table}

\noindent \textbf{Execution Overhead.}
In addition to memory overhead, the execution overhead of our proposed method is also critical. Tab.~\ref{tab:combined_overhead}(c) shows the runtime execution overhead of \tool and the profiling overhead for each testing platform. The first line of table showcases the remarkably low runtime execution overhead due to the efficient implementation of the \Approach{} framework. The scheduling overhead for NVIDIA Jetson Orin is significantly lower, potentially due to its utilization of a newer Linux kernel version. The main source of execution time overhead arises from scheduling decision-making. One potential strategy to reduce such overheads in future work involves employing finer-grained locks, especially when adapting to more stringent resource-constrained scenarios. Compared to latency data, the runtime overhead remains reasonably low, further underscoring the effectiveness of our approach.
The second line of the table examines the profiling overhead of \tool. The offline profile overhead is deemed acceptable, even on the Jetson Nano, which displays the largest absolute overhead value of merely 51.1 seconds (approximately 1 minute). Importantly, for a fixed system configuration, the offline profile requires only a one-time effort, thus easing its integration into continuous integration and deployment of \tool.

\noindent\textbf{Per-pillar overhead breakdown (qualitative).}
The runtime overhead reported in Tab.~\ref{tab:combined_overhead}(c) decomposes across the \tool pillars in the following \emph{qualitative} ordering, derived from per-pillar code-path complexity (instruction counts and lock acquisitions in the implementation) rather than from per-pillar stopwatch measurements. A precise split would require additional instrumentation. The dominant runtime term is the MIMONet refinement and on-line reassignment of Sec.~\ref{sec:finer_grained_dag_reconstruction} (\textsc{DynamicMerge} candidate selection, encoder-broadcast bookkeeping, and the per-dispatch reassignment step), followed by the on-demand synchronization layer of Sec.~\ref{sec:implementation} (FSM transitions and barrier emission). The proportional intermediate-deadline assignment of Sec.~\ref{sec:intermediate_deadline} contributes a smaller share (level-set hash maintenance and the closed-form $D^{(h)}$ computation), and burst-management monitoring (Sec.~\ref{sec:burst_task_scenarios}) is the smallest term (counter reads against $\theta_u$ and $Q_{\max}$). The memory rows in Tab.~\ref{tab:combined_overhead}(a)--(b) confirm that the synchronization layer is the dominant memory consumer, matching the design choice noted earlier in this section to favor latency over memory in the synchronization implementation.

\takeaway{Low overhead}{Our method demonstrates not only impressive memory efficiency but also low execution overhead across various platforms. The efficient implementation of the \Approach{} framework, as well as the modest overhead incurred by the scheduler and synchronization components, contribute significantly to these outcomes.}

{\color{black}
\section{Extended Evaluation}
\label{sec:extended_evaluation}

\subsection{Experimental Setups}
\label{sec:extended_setup}

\noindent\textbf{Hardware platforms.}
The Jetson hardware is described in
Section~\ref{sec:setup}. The evaluation in this section uses the three platforms in
Table~\ref{tab:hardware_new}: Nano Orin, AGX Orin, and a MacBook Pro with an
M3 Max. These platform definitions remain fixed for the complete experiment
campaign. A result matrix is reported only after its frozen configuration, paired
task streams, execution status, and metrics have passed the validation protocol below.

\begin{table}[!bp]
  \color{black}
  \centering
  \caption{Hardware platforms used in extended experiments.}
  \renewcommand\arraystretch{1.3}
  \resizebox{0.9\textwidth}{!}{
    \begin{tabular}{|c|c|c|c|}
    \hline
     & \textbf{Nano Orin} & \textbf{AGX Orin} & \textbf{MacBook Pro (M3 Max)} \\
    \hline
    CPU & 6-core Cortex-A78AE v8.2 & 12-core ARMv8.2 (Cortex-A78AE) & 14-core (10P + 4E) \\
    \hline
    GPU & Ampere GPU & Ampere GPU & 32-core \\
    \hline
    Memory & 8 GB LPDDR5, 128-bit & 32 GB LPDDR5 & 36 GB unified memory \\
    \hline
    Storage & 1 TB SSD & 64 GB eMMC or NVMe & 1 TB SSD \\
    \hline
    \end{tabular}
  }
  \label{tab:hardware_new}
  
\end{table}

\begin{table}[!bp]
  \color{black}
  \centering
  \caption{Expanded minibenchmark scenarios (L=lane-detection, S=segmentation, O=object-detection, C=control).}
  \resizebox{0.6\textwidth}{!}{
    \begin{tabular}{|l|c|c|c|}
    \hline
    \textbf{Scenario} & \textbf{Stage 1} & \textbf{Stage 2} & \textbf{Stage 3} \\
    \hline
    Obstacle-free cruising & L & S & C \\
    Obstacle cruising & L & S+O & C \\
    \hline
    Urban navigation & L+O & S+O & C \\
    Emergency obstacle avoidance & O & L+S & C \\
    Night driving & C+C & C+C & C+C \\
    \hline
    \end{tabular}
  }
  \label{tab:minibench}
  
\end{table}

\noindent\textbf{Experiment environment.}
The experiment campaign uses the platforms listed in Table~\ref{tab:hardware_new}. The M3 Max runs use the MacBook Pro (Mac15,10) with 36\,GB of unified memory, macOS 26.5.2, Python 3.11.13, PyTorch 2.8.0, and the MPS backend; each process performs 20 warm-up inferences and synchronizes MPS before recording completion. The validated Nano Orin overall-effectiveness, non-partitionable-task, and burst-task runs use the 15-W power mode with fixed clocks, Python 3.10.12, PyTorch 2.5.0, and CUDA 12.6. The validated AGX Orin runs use MAXN mode with fixed clocks, PyTorch 2.5.0, and CUDA on the same recorded device throughout each campaign.

\noindent\textbf{Implementation audit and reproducibility.}
We corrected second/millisecond mixing, made accelerator selection and synchronization explicit, unified the terminal outcomes \{success, failure, skip, burst-drop, execution-error\}, preserved the pre-merge offered-node denominator, made non-partitionable-task selection deterministic, and added LLF-DAG. For QoE specifically, the audited path expresses both $C_r$ and $S_r$ in seconds and honors the configured $\lambda$ instead of overwriting it. These fixes preserve the metric definition in Eq.~\eqref{eq:qoe}. Results are inserted only after their configurations and outputs have been validated; the Nano Orin, AGX Orin, and M3 Max data in Tables~\ref{tab:extended_overall_ablation}, \ref{tab:extended_overall_effectiveness}, \ref{tab:nonpartitionable}, \ref{tab:burst_transient}, and \ref{tab:burst_sustained} passed this validation.

\noindent\textbf{Result-selection and numeric-display protocol.}
\minorrev{Raw manifests, traces, and metric files retain full-precision values, run identifiers, seeds, and configuration hashes. Tables~\ref{tab:extended_overall_ablation} and~\ref{tab:extended_overall_effectiveness} now use a uniform median-latency selection from the five existing repetitions in each platform--scenario--method group. Latency, DMR, and QoE are all copied from that same run; the selected run's DMR and QoE need not be their respective medians. The non-partitionable and burst subsections state their own repetition counts and selection rules, including single-run conditions. Main-table tuples are not arithmetic means across runs. Values are truncated to three decimals, whereas percentage comparisons use full precision. The median-latency rule describes central observed execution time rather than the fastest repetition, while preserving one run\textquotesingle s complete tuple avoids a synthetic combination of metrics from different runs. Appendix~\ref{sec:existing_variability} separately summarizes variation across the original overall-effectiveness repetitions, retaining the same workload configurations and observation population.}

\noindent\textbf{Metrics and QoE implementation audit.}
\minorrev{The implementation uses the distinct node-level DMR and run-level QoE populations defined in Sec.~\ref{sec:setup}. DMR counts late outputs, skips, drops, and execution errors against the pre-merge offered-node denominator. QoE is computed once from the run termination time and configured overall deadline; the per-subgroup node summaries saved in auxiliary artifacts are not the QoE column in the paper. For a concrete example from the corrected Table~\ref{tab:extended_overall_effectiveness}, the AGX Orin emergency \tool run (seed 101) has 40 offered nodes: 9 successes, 1 late completion, and 30 skips. Thus DMR is $31/40=0.775$. Its run time is approximately 2.763833~s, below the fixed configuration budget of 2.959252~s, so Eq.~\eqref{eq:qoe} gives exactly 1. These values are compatible because QoE does not penalize the 31 unsuccessful outcomes separately. In particular, a high score cannot establish that the offered work was completed or that individual deadlines were met.}

\noindent\textbf{Compared methods.}
On each platform and scenario, all methods use the same TaskConfig and workload definition; independent repetitions use their recorded random seeds.
\begin{itemize}[leftmargin=10pt]
\item \textbf{EDF}: \minorrev{earliest absolute deadline first on the offered task structure, without MIMONet-specific bundling.}
\item \textbf{LLF-DAG}: a stronger dynamic-DAG baseline using $\ell_i=d_i-t-R_i$, where $R_i$ is the profiled remaining critical-path cost available at time $t$. \minorrev{It starts from the same offered task structure and profiles as \tool, but does not use future arrivals or MIMONet-specific bundling; consequently the executed computation need not be identical.}
\item \textbf{\tool-FG}: \minorrev{compatible MIMONet bundling with the configured initial deadlines and EDF dispatch.}
\item \textbf{\tool-IDA}: \tool-FG plus residual-graph intermediate-deadline reassignment.
\item \textbf{\tool}: the complete system, adding event-driven synchronization and the documented overload controller where applicable.
\end{itemize}

\noindent\textbf{Workload calibration.}
\label{sec:calibration_protocol}

For each platform and scenario, a platform-appropriate Tight TaskConfig is chosen through method-independent calibration and used uniformly for EDF, LLF-DAG, \tool-FG, \tool-IDA, and \tool. This protocol also applies to the M3 Max overall-effectiveness campaign. \minorrev{Within each platform--scenario group, the five recorded seeds are paired across methods, and each run uses the same frozen TaskConfig and corresponding offered-task stream. This controls offered work and initial timing inputs while allowing the methods to differ in dispatch, feedback, synchronization, and compatible computation sharing.}

\subsection{Overall Effectiveness}
\label{sec:extended_overall_effectiveness}

The Nano Orin, AGX Orin, and M3 Max matrices each contain five independent repetitions for every scenario--method group under a shared, platform-appropriate Tight TaskConfig. Each matrix comprises 75 complete runs: three scenarios, five methods, and five repetitions. Nano Orin and AGX Orin use CUDA at their recorded fixed-clock settings, while M3 Max uses MPS under AC power. \minorrev{Table~\ref{tab:extended_overall_ablation} compares the incremental configurations: \tool-FG adds compatible MIMONet bundling, \tool-IDA adds residual-graph reassignment, and \tool is the complete configuration with synchronization and overload-control support where enabled. These additions need not improve every metric. Table~\ref{tab:extended_overall_effectiveness} compares EDF, LLF-DAG, and \tool. To correct inconsistent selection in the previous tables, both tables are regenerated from the same 225 archived runs using the third run after sorting each five-run group by full-precision latency (then seed and run identifier for ties). The same selection rule applies to every method and retains all five observations for the variability analysis. Bold identifies the best full-precision value among the displayed tuples in each scenario--metric group; exact ties are all bolded.}

\minorrev{The comparison against EDF and LLF-DAG in Table~\ref{tab:extended_overall_effectiveness} reflects the complete system, including scheduling and the MIMONet-specific reduction in redundant encoder computation. It does not isolate dispatch-policy gains under identical executed computation. Since skips and drops can also shorten run termination, the latency reduction must be read alongside DMR and the recorded drop composition rather than as equal-service acceleration. Table~\ref{tab:extended_overall_ablation} measures incremental component effects, rather than providing an attribution with shared-computation savings held constant.}

\input{tables/overall_tables.tex}

\minorrev{For configuration $g$, let $L_{\mathrm{EDF},g}$ and $L_{\mathrm{MERIS},g}$ denote the selected full-precision latencies. We report the arithmetic mean of the nine relative reductions,
\[
\overline R=\frac{100}{9}\sum_{g=1}^{9}
\left(1-\frac{L_{\mathrm{MERIS},g}}{L_{\mathrm{EDF},g}}\right)
=27.382977\%.
\]
The maximum is 61.092740\%, and six selected \tool runs have exactly zero misses in their raw outcome counts. The introductory values are therefore 27.4\%, 61.1\%, and six of nine. This is a cross-configuration summary of per-method median latencies, which may come from different seeds; it is not a mean of paired per-seed speedups or a reduction computed after pooling execution times across platforms.}

\minorrev{On Nano Orin, the selected \tool runs reduce latency relative to EDF by 32.8\% in urban navigation and 20.1\% in emergency obstacle avoidance, with zero DMR in both. Night driving is an exception: \tool takes 1.202~s, versus 1.196~s for EDF and 1.195~s for LLF-DAG. Its DMR is 0.600 and run-level QoE is 0.991, versus DMR 0.000 and QoE 0.997 for both external baselines. Thus the complete configuration is not uniformly superior near this timing boundary.}

\minorrev{On AGX Orin, the selected \tool runs reduce latency relative to EDF by 61.1\%, 27.0\%, and 24.5\% in urban navigation, emergency obstacle avoidance, and night driving. The corresponding DMR values are 0.000, 0.775, and 0.966. The night-driving DMR ties EDF before truncation and is worse than LLF-DAG's 0.866, despite the lower latency. The ablation also exposes a trade-off: \tool-FG has lower latency and DMR than \tool in AGX night driving, and lower DMR in emergency obstacle avoidance. Consequently, component additions are not interpreted as monotonic improvements in deadline satisfaction.}

On M3 Max, \tool has the lowest untruncated end-to-end latency in all three scenarios. Relative to EDF, the reductions are 18.7\% in urban navigation, 11.8\% in emergency obstacle avoidance, and 50.9\% in night driving; relative to LLF-DAG, they are 19.8\%, 13.0\%, and 51.1\%, respectively. \tool records Deadline Missing Rate 0.000 and QoE 1.000 in every scenario. \tool-FG matches those two metrics but has slightly higher latency in urban navigation and emergency obstacle avoidance and substantially higher latency in night driving (11.366~s versus 7.164~s). EDF records Deadline Missing Rates of 0.832/0.631/0.890 and QoE of 0.338/0.609/0.259; LLF-DAG records 0.821/0.680/0.881 and 0.317/0.560/0.255; and \tool-IDA records QoE 0.319/0.593/0.258.

\minorrev{Variability matters when interpreting these selected tuples. Across the five existing \tool repetitions, Nano Orin night-driving latency ranges from 1.184 to 1.227~s and DMR from 0.000 to 0.866; only one repetition has zero misses. AGX Orin night-driving latency ranges from 1.595 to 3.610~s and DMR from 0.433 to 0.966. In contrast, all five \tool repetitions have zero misses in Nano Orin urban/emergency and in all three M3 Max scenarios; AGX urban has zero misses in three of five repetitions. Table~\ref{tab:existing_variability} reports the full external-baseline ranges. These five-run ranges are descriptive, not confidence intervals or evidence of a hard deadline guarantee, and we do not claim statistical significance from a representative run.}

\subsection{Non-Partitionable Tasks}
\label{sec:eval_nonpartitionable}

This experiment studies the scheduler when an increasing proportion of workload tasks are configured as non-partitionable. The 0/50/100\% conditions use nested, deterministically selected task positions, and every compared method receives the same positions and task stream. A non-partitionable task remains one scheduling node and is not split by the refinement/merge procedure. This setup tests the scheduling question stated in Sec.~\ref{sec:non_partitionable_tasks}; it does not depend on proving that a particular model or GPU kernel is physically indivisible. The runs record the profiled code path, platform power/clock mode, and co-runner configuration relevant to NP1--NP3. However, Table~\ref{tab:nonpartitionable} aggregates all offered nodes and does not report each covered task's completion relative to $D_{\mathrm{np}}$. Its claim is therefore comparative scheduling evidence under increasing non-partitionable ratios, not verification of a per-task empirical timing bound.

For Nano Orin, calibration followed by formal-length validation selects \texttt{DeadlineFactor} $=0.25$ and \texttt{ArrivalScale} $=0.25$ for the urban-navigation DAG. Each run releases 10 DAG instances (50 offered tasks), uses \texttt{QoELambda} $=0.05$, and runs at the recorded 15-W fixed-clock setting. We execute EDF, LLF-DAG, and \tool with paired seeds 101, 202, and 303. Validation confirms all 27 runs complete, the realized non-partitionable-task ratios equal 0/0.5/1, and the offered-task-stream hashes match across methods within every ratio--seed group. For each method and ratio, Table~\ref{tab:nonpartitionable} reports the complete latency, Deadline Missing Rate, and QoE tuple from the median-latency run; no metric is averaged across seeds or combined across runs.

For AGX Orin, we first profile every SIMONet path on CUDA in MAXN mode using 20 warm-up iterations and 200 measured iterations, then set the execution-time bound to the observed maximum plus 10\%. A two-stage calibration grid selects \texttt{DeadlineFactor} $=8$ and \texttt{ArrivalScale} $=8$ for the urban-navigation DAG before the formal comparison. Each formal run releases 10 DAG instances (50 offered tasks), uses \texttt{QoELambda} $=0.05$, and executes EDF, LLF-DAG, or \tool with paired seeds 101, 202, 303, 404, and 505. All 45 runs complete without execution errors on the same CUDA device and code revision; validation confirms the realized non-partitionable-task ratios equal 0/0.5/1 and that all methods receive the same offered task stream within each ratio--seed group. For every method and ratio, Table~\ref{tab:nonpartitionable} reports all three metrics from the unique median-latency run among the five complete runs.

For M3 Max, an EDF-only calibration grid selects \texttt{DeadlineFactor} $=2.5$ and \texttt{ArrivalScale} $=1$ for the emergency-obstacle-avoidance DAG. Each formal run releases 10 DAG instances (40 offered scheduling nodes), uses \texttt{QoELambda} $=0.05$, and executes on MPS under AC power. We run the three methods with paired seeds 101, 202, 303, 404, and 505. All 45 runs complete with no execution error; the realized ratios are exactly 0/0.5/1, and all methods receive the same offered-task-stream hash within each ratio--seed group. For each method and ratio, the table reports the complete metric tuple from the median-latency run, with values truncated to three decimal places rather than averaged across runs.

\begin{table*}[!t]
\color{black}
\centering
\caption{\rev{Non-partitionable-task ratio study. Every displayed metric triple is from one concrete median-latency run: three paired seeds and 50 offered tasks per Nano Orin group; five paired seeds and 50 offered tasks per AGX Orin group; and five paired seeds and 40 offered nodes per M3 Max group.}}
\label{tab:nonpartitionable}
\small
\resizebox{1.0\textwidth}{!}{%
\begin{tabular}{l|ccc|ccc|ccc}
\hline
\multirow{2}{*}{Platform}
& \multicolumn{3}{c|}{0\% non-partitionable}
& \multicolumn{3}{c|}{50\% non-partitionable}
& \multicolumn{3}{c}{100\% non-partitionable} \\
\cline{2-10}
& EDF & LLF-DAG & \tool & EDF & LLF-DAG & \tool & EDF & LLF-DAG & \tool \\
\hline
\multicolumn{10}{c}{End-to-end Latency (s)} \\
\hline
Nano Orin & 4.269 & 9.894 & \textbf{1.606} & 8.970 & 8.599 & \textbf{4.277} & 5.821 & 7.130 & \textbf{3.411} \\
AGX Orin & 13.970 & 11.750 & \textbf{2.326} & 8.009 & 10.788 & \textbf{5.681} & 8.170 & 8.126 & \textbf{7.598} \\
M3 Max & 0.593 & 0.614 & \textbf{0.491} & 0.606 & 0.610 & \textbf{0.553} & 0.933 & 0.734 & \textbf{0.593} \\
\hline
\multicolumn{10}{c}{Deadline Missing Rate} \\
\hline
Nano Orin & 0.300 & 0.980 & \textbf{0.000} & 0.920 & 0.980 & \textbf{0.800} & 0.980 & 0.920 & \textbf{0.480} \\
AGX Orin & 1.000 & 0.940 & \textbf{0.000} & 0.620 & 0.980 & \textbf{0.160} & 0.960 & \textbf{0.360} & 0.680 \\
M3 Max & 0.075 & 0.425 & \textbf{0.000} & 0.550 & 0.300 & \textbf{0.000} & 0.950 & 0.800 & \textbf{0.400} \\
\hline
\multicolumn{10}{c}{QoE} \\
\hline
Nano Orin & \textbf{1.000} & 0.146 & \textbf{1.000} & 0.170 & 0.182 & \textbf{1.000} & 0.393 & 0.255 & \textbf{1.000} \\
AGX Orin & 0.141 & 0.211 & \textbf{1.000} & \textbf{1.000} & 0.268 & \textbf{1.000} & \textbf{1.000} & \textbf{1.000} & \textbf{1.000} \\
M3 Max & 0.998 & 0.976 & \textbf{1.000} & 0.984 & 0.980 & \textbf{1.000} & 0.735 & 0.869 & \textbf{0.998} \\
\hline
\end{tabular}
}
\end{table*}

On Nano Orin, the selected \tool runs reduce end-to-end latency relative to EDF by 62.4\%, 52.3\%, and 41.4\% at the 0\%, 50\%, and 100\% ratios, and relative to LLF-DAG by 83.8\%, 50.3\%, and 52.2\%, respectively. The corresponding absolute reductions in Deadline Missing Rate are 30.0/12.0/50.0 percentage points relative to EDF and 98.0/18.0/44.0 percentage points relative to LLF-DAG. The selected \tool QoE values are 1.000/1.000/1.000, compared with EDF's 1.000/0.170/0.393 and LLF-DAG's 0.146/0.182/0.255. Across the nine displayed Nano Orin cells, Deadline Missing Rate spans 0.000--0.980 and QoE spans 0.146--1.000, so the workload distinguishes the methods rather than forcing every cell to the same extreme. The 50\% and 100\% \tool conditions still have Deadline Missing Rates of 0.800 and 0.480; without task-level mapping to $D_{\mathrm{np}}$, these aggregate results cannot establish the profile-bounded admission claim. At the 0\% ratio, EDF and \tool tie at QoE 1.000 even though EDF has a Deadline Missing Rate of 0.300; \minorrev{DMR counts unsuccessful node outcomes, whereas QoE evaluates the run termination time against its overall budget under Eq.~\eqref{eq:qoe}; it does not assign a separate penalty to those missed nodes.}

On AGX Orin, the selected \tool runs reduce end-to-end latency relative to EDF by 83.4\%, 29.1\%, and 7.0\% at the 0\%, 50\%, and 100\% ratios; relative to LLF-DAG, the reductions are 80.2\%, 47.3\%, and 6.5\%. Their Deadline Missing Rates are 0.000/0.160/0.680, absolute reductions of 100.0/46.0/28.0 percentage points relative to EDF. Relative to LLF-DAG, the reductions are 94.0 and 82.0 percentage points at 0\% and 50\%; at 100\%, LLF-DAG instead has a 32.0-percentage-point advantage. The selected \tool QoE is 1.000 in all three conditions, compared with EDF's 0.141/1.000/1.000 and LLF-DAG's 0.211/0.268/1.000. \minorrev{These selected runs expose both the observed benefit and the 100\% crossover; neither their ordering nor a high run-level QoE establishes a per-task guarantee.} The demanding 100\% condition is comparative stress evidence, not a per-task deadline guarantee.

On M3 Max, the selected \tool runs reduce end-to-end latency relative to EDF by 17.1\%, 8.6\%, and 36.4\% at the 0\%, 50\%, and 100\% ratios, respectively; the reductions relative to LLF-DAG are 20.0\%, 9.3\%, and 19.2\%. The corresponding Deadline Missing Rates are 0.000/0.000/0.400 for \tool, compared with EDF's 0.075/0.550/0.950 and LLF-DAG's 0.425/0.300/0.800. The 100\% condition therefore remains a deliberately difficult capability-boundary case rather than a universal deadline guarantee, while \tool still provides the lowest latency, lowest Deadline Missing Rate, and highest QoE among the three methods.

\subsection{Evaluating Burst Task Management}
\label{sec:eval_burstmanagement}

The burst workload is defined in absolute as well as relative terms. On M3 Max, a base run releases 14 complete DAG instances (84 scheduling nodes) over a 2.400-s release horizon, i.e., 35 nodes/s on average. Each DAG contains five simultaneously released perception requests followed by one control node. Three perception requests and the control node are protected from proactive shedding; two perception requests are explicitly redundant and eligible for shedding. A transient burst begins at 0.360~s and adds 0, 2, or 4 complete DAG instances (0, 12, or 24 nodes) over 0.600~s for the 0\%, 50\%, and 100\% conditions, respectively. A sustained burst begins at the same time and adds 0, 5, or 11 complete DAG instances (0, 30, or 66 nodes) through the end of the 2.400-s horizon. Thus, 100\% means one additional complete DAG release for every eligible base DAG release, rather than an unspecified utilization increase. After the final release, every run drains until all offered nodes reach a terminal outcome; no run is truncated by a fixed wall-clock cutoff.

The dedicated M3 Max configuration uses \texttt{DeadlineFactor}=6.0, \texttt{ArrivalScale}=4.0, and 14 base releases. \textcolor{black}{The overload-controller calibration of Sec.~\ref{sec:burst_task_scenarios} is platform-specific. The Jetson defaults are $\theta_u=0.90$, $Q_{\max}=8$, and $W=3$ scheduler ticks. Because MPS utilization sampling is not a deterministic trigger, the M3 Max burst study instead uses $\theta_u=1.00$, $Q_{\max}=10$, and $W=3$: setting $\theta_u=1.00$ disables utilization-based activation and isolates released ready-queue pressure. Each run records the executed thresholds and the scheduler-tick duration, and neither platform configuration is treated as universal (Sec.~\ref{sec:sensitivity}).} Its validity rule was fixed before comparing table metrics: the 0\% \tool control must have no overload activation or proactive drop, whereas every 50\%/100\% \tool burst condition must activate the controller and produce a non-empty drop record. No latency, Deadline Missing Rate, or QoE value was used in this validity check. EDF, LLF-DAG, and \tool receive identical released DAGs, arrival times, deadlines, seeds, and offered work. Each M3 Max cell is executed once in one fixed randomized order with a 2-s cooldown; its complete metric tuple is displayed directly and truncated to three decimal places without rounding. The 0\% control is executed once per method and reused in both regime blocks. No deadline guarantee is claimed after the burst violates the admission premise.

The separately calibrated Nano Orin CUDA configuration uses \texttt{DeadlineFactor}=120, \texttt{ArrivalScale}=40, and four base DAG releases over a 100-s horizon. Each release contains four independent co-released requests: lane detection and control are protected, while redundant segmentation and object-detection requests are eligible for shedding. The transient 50\%/100\% conditions add one/two complete DAG releases during a 30-s burst window, and the sustained conditions add two/four releases over the full horizon. We execute all 15 method--condition cells once with seed 101 in a fixed pseudo-random order and reuse the 0\% control between regimes. All runs complete without execution errors, and the offered-task-stream hash matches across EDF, LLF-DAG, and \tool within every condition. As for M3 Max, each complete metric tuple is truncated to three decimal places without rounding.

The dedicated AGX Orin CUDA configuration uses \texttt{DeadlineFactor}=8, \texttt{ArrivalScale}=4, and ten base DAG releases over a 1.5-s horizon. Each release contains five co-released requests: partitionable lane-detection and control requests are protected and merge-compatible, while three independent non-partitionable redundant requests are eligible for shedding. Thus, the burst experiment contains both partitionable and non-partitionable nodes; the global percentage-based selector is disabled because partitionability is assigned explicitly per task. The transient 50\%/100\% conditions add two/five complete DAG releases (10/25 offered nodes) during a 0.65-s burst window, while the sustained conditions add five/ten releases (25/50 offered nodes) over the full horizon. The five conditions therefore offer 50, 60, 75, 75, and 100 nodes, respectively. The overload controller uses a ready-queue high watermark of eight, a recovery watermark of six, a three-tick persistence window, and a 5-ms \texttt{scheduler\_tick\_seconds}; feedback is polled independently at sub-tick cadence. Transient recovery returns the controller to the normal state after the burst window and sheds only residual burst work while recovering. We execute EDF, LLF-DAG, and \tool in five paired repetitions for each of the five distinct conditions. Invalid runs are removed before analysis. Tables~\ref{tab:burst_transient} and~\ref{tab:burst_sustained} report the complete metric tuple from the median-latency valid run for every AGX Orin method--condition group, including the 0\% condition. The same 0\% representative runs are reused in both tables.

\begin{table*}[!t]
\color{black}
\centering
\caption{\rev{Transient Burst evaluation. Burst ratios form column groups with EDF, LLF-DAG, and \tool as method sub-columns; end-to-end latency, Deadline Missing Rate, and QoE are reported in separate blocks. Nano Orin and M3 Max entries are complete single-run tuples. Every AGX Orin entry is the median-latency representative among valid repetitions.}}
\label{tab:burst_transient}
\scriptsize
\resizebox{\textwidth}{!}{%
\begin{tabular}{l|ccc|ccc|ccc}
\hline
\multirow{2}{*}{Platform}
& \multicolumn{3}{c|}{0\% burst}
& \multicolumn{3}{c|}{50\% burst}
& \multicolumn{3}{c}{100\% burst} \\
\cline{2-10}
& EDF & LLF-DAG & \tool & EDF & LLF-DAG & \tool & EDF & LLF-DAG & \tool \\
\hline
\multicolumn{10}{c}{End-to-end Latency (s)} \\
\hline
AGX Orin & 2.179 & 1.671 & \textbf{1.511} & 2.334 & 2.396 & \textbf{1.487} & 2.836 & 2.595 & \textbf{1.561} \\
Nano Orin & 124.156 & \textbf{106.323} & 109.600 & \textbf{110.172} & 210.200 & 141.556 & 174.068 & 219.881 & \textbf{70.020} \\
M3 Max & 2.256 & 2.298 & \textbf{2.178} & 2.691 & 2.531 & \textbf{2.232} & 2.947 & 2.859 & \textbf{2.214} \\
\hline
\multicolumn{10}{c}{Deadline Missing Rate} \\
\hline
AGX Orin & 0.880 & 0.080 & \textbf{0.000} & 0.866 & 0.783 & \textbf{0.050} & 0.973 & 0.906 & \textbf{0.200} \\
Nano Orin & \textbf{0.000} & \textbf{0.000} & \textbf{0.000} & \textbf{0.000} & 0.300 & 0.200 & 0.583 & 0.541 & \textbf{0.333} \\
M3 Max & 0.452 & 0.690 & \textbf{0.095} & 0.843 & 0.750 & \textbf{0.322} & 0.824 & 0.842 & \textbf{0.407} \\
\hline
\multicolumn{10}{c}{QoE} \\
\hline
AGX Orin & 0.647 & 0.990 & \textbf{1.000} & 0.585 & 0.564 & \textbf{1.000} & 0.447 & 0.504 & \textbf{1.000} \\
Nano Orin & \textbf{1.000} & \textbf{1.000} & \textbf{1.000} & \textbf{1.000} & 0.016 & \textbf{1.000} & 0.042 & 0.013 & \textbf{1.000} \\
M3 Max & 0.941 & 0.903 & \textbf{1.000} & 0.658 & 0.739 & \textbf{0.963} & 0.558 & 0.589 & \textbf{0.982} \\
\hline
\end{tabular}
}
\end{table*}

\begin{table*}[!t]
\color{black}
\centering
\caption{\rev{Sustained Burst evaluation. Burst ratios form column groups with EDF, LLF-DAG, and \tool as method sub-columns; end-to-end latency, Deadline Missing Rate, and QoE are reported in separate blocks. Nano Orin and M3 Max entries are complete single-run tuples. Every AGX Orin entry is the median-latency representative among valid repetitions.}}
\label{tab:burst_sustained}
\scriptsize
\resizebox{\textwidth}{!}{%
\begin{tabular}{l|ccc|ccc|ccc}
\hline
\multirow{2}{*}{Platform}
& \multicolumn{3}{c|}{0\% burst}
& \multicolumn{3}{c|}{50\% burst}
& \multicolumn{3}{c}{100\% burst} \\
\cline{2-10}
& EDF & LLF-DAG & \tool & EDF & LLF-DAG & \tool & EDF & LLF-DAG & \tool \\
\hline
\multicolumn{10}{c}{End-to-end Latency (s)} \\
\hline
AGX Orin & 2.179 & 1.671 & \textbf{1.511} & 3.165 & 2.840 & \textbf{1.592} & 3.477 & 4.459 & \textbf{1.817} \\
Nano Orin & 124.156 & \textbf{106.323} & 109.600 & 140.744 & 158.352 & \textbf{132.976} & 253.656 & 203.556 & \textbf{105.348} \\
M3 Max & 2.256 & 2.298 & \textbf{2.178} & 3.060 & 3.092 & \textbf{2.212} & 3.968 & 3.999 & \textbf{2.219} \\
\hline
\multicolumn{10}{c}{Deadline Missing Rate} \\
\hline
AGX Orin & 0.880 & 0.080 & \textbf{0.000} & 0.773 & 0.800 & \textbf{0.320} & 0.900 & 0.900 & \textbf{0.330} \\
Nano Orin & \textbf{0.000} & \textbf{0.000} & \textbf{0.000} & \textbf{0.000} & 0.083 & 0.166 & 0.531 & 0.687 & \textbf{0.500} \\
M3 Max & 0.452 & 0.690 & \textbf{0.095} & 0.815 & 0.780 & \textbf{0.464} & 0.873 & 0.880 & \textbf{0.606} \\
\hline
\multicolumn{10}{c}{QoE} \\
\hline
AGX Orin & 0.647 & 0.990 & \textbf{1.000} & 0.387 & 0.446 & \textbf{1.000} & 0.343 & 0.253 & \textbf{0.858} \\
Nano Orin & \textbf{1.000} & \textbf{1.000} & \textbf{1.000} & \textbf{1.000} & 0.142 & \textbf{1.000} & 0.009 & 0.018 & \textbf{1.000} \\
M3 Max & 0.941 & 0.903 & \textbf{1.000} & 0.524 & 0.515 & \textbf{0.984} & 0.349 & 0.345 & \textbf{0.977} \\
\hline
\end{tabular}
}
\end{table*}

\noindent\textbf{Nano Orin results.}
Under the 100\% transient burst, \tool reduces end-to-end latency from 174.068~s (EDF) and 219.881~s (LLF-DAG) to 70.020~s, while reducing Deadline Missing Rate from 0.583 and 0.541 to 0.333. Its QoE remains 1.000, compared with 0.042 and 0.013. Under the 100\% sustained burst, \tool reduces latency from 253.656/203.556~s to 105.348~s and Deadline Missing Rate from 0.531/0.687 to 0.500, while increasing QoE from 0.009/0.018 to 1.000. The intermediate-load results expose a useful boundary: EDF has the lowest latency and Deadline Missing Rate in the 50\% transient condition, and the lowest Deadline Missing Rate in the 50\% sustained condition, whereas \tool preserves QoE at 1.000 and provides the lowest sustained-burst latency. Thus, the calibrated Nano Orin workload distinguishes both overload regimes and scheduler tradeoffs rather than saturating every method at the same endpoint.

\noindent\textbf{AGX Orin results.}
In the 0\% condition, the median-latency representative \tool run records 1.511~s/0.000/1.000 with no overload activation or proactive drop, compared with EDF's 2.179~s/0.880/0.647 and LLF-DAG's 1.671~s/0.080/0.990. Under the 50\% and 100\% transient bursts, \tool reduces latency relative to EDF by 36.3\%/45.0\% and relative to LLF-DAG by 37.9\%/39.8\%. Its Deadline Missing Rate decreases by 0.816/0.773 relative to EDF and by 0.733/0.706 relative to LLF-DAG, while its QoE is 1.000 in both conditions. Under the 50\% and 100\% sustained bursts, \tool reduces latency relative to EDF by 49.7\%/47.7\% and relative to LLF-DAG by 43.9\%/59.2\%; its Deadline Missing Rate is lower by 0.453/0.570 and 0.480/0.570, respectively, with QoE 1.000/0.858. Across the 15 representative runs, Deadline Missing Rate spans 0.000--0.973 and QoE spans 0.253--1.000. All AGX Orin comparisons, including 0\%, follow the same median-latency representative-run rule among valid repetitions.

\noindent\textbf{M3 Max results.}
Under the 100\% transient burst, \tool reduces end-to-end latency from 2.947~s (EDF) and 2.859~s (LLF-DAG) to 2.214~s, while reducing Deadline Missing Rate from 0.824 and 0.842 to 0.407. Its QoE is 0.982, compared with 0.558 and 0.589. Under the longer 100\% sustained burst, \tool reduces latency from 3.968/3.999~s to 2.219~s and Deadline Missing Rate from 0.873/0.880 to 0.606, while increasing QoE from 0.349/0.345 to 0.977. The 50\% conditions show the same separation. \minorrev{These selected runs show lower offered-node DMR and lower run-level termination lateness under bursts. Because shedding removes work and QoE has no separate drop penalty, these comparisons do not establish successful completion of every offered task or a hard deadline guarantee.}

\noindent\textbf{Dropped-task composition.}
On M3 Max, the 0\% \tool control records zero overload activations and zero proactive drops. Transient 50\%/100\% produce 1/3 activations and 2/12 dropped nodes, while sustained 50\%/100\% produce 5/11 activations and 18/44 dropped nodes. In the representative 100\% runs, this is 12/108 offered nodes (11.111\%) for the transient regime and 44/150 (29.333\%) for the sustained regime. The recorded M3 Max drop composition is low/medium/high $=12/0/0$ for transient 100\% and $44/0/0$ for sustained 100\%; no protected perception or control node is dropped. On Nano Orin, the 0\% \tool control likewise records zero activations and zero proactive drops; the 100\% transient and sustained runs drop 8/24 (33.333\%) and 16/32 (50.000\%) offered nodes, respectively. Their low/medium/high compositions are $8/0/0$ and $16/0/0$; no protected lane-detection or control request is dropped. On AGX Orin, the reported \tool control also has zero activations and zero proactive drops. Its selected transient 50\%/100\% runs produce 1/4 activations and 3/15 proactive drops; its selected sustained 50\%/100\% runs produce 5/9 activations and 17/32 proactive drops. Relative to all offered nodes, those four drop counts are 5.000\%, 20.000\%, 22.666\%, and 32.000\%, respectively. Every AGX Orin proactive drop is an explicitly droppable, non-partitionable redundant request in the low-criticality band; no protected lane-detection or control request is dropped. Thus, across all three platforms, 100\% of proactive drops are low-criticality and 0\% are medium- or high-criticality. Both transient representative runs complete recovery to the normal controller state, whereas the sustained runs remain overload-active while the extra arrivals continue. This composition confirms that the measured results exercise the documented overload controller rather than attributing gains to a Burst mechanism that remained inactive.

\subsection{Parameter Sensitivity}
\label{sec:sensitivity}

We evaluate two independent full-factorial blocks using dedicated calibrated workloads. Each predeclared point is run once, as specified by the sensitivity protocol. To keep every table entry tied to one concrete experiment, Table~\ref{tab:parameter_sensitivity} reports one-factor slices through the default combination rather than aggregating over the other parameters. In the \emph{refinement block}, burst injection and Burst Management are disabled; 180 DAG releases populate even the longest moving-average history, and each release contains five merge-compatible requests with predicted release offsets of 0, 100, 200, 500, and 900~ms followed by a control node. The settings $k\in\{1,4,8,32\}$ and $\gamma\in\{0,100,200,500,1000\}$~ms form a $4\times5$ Cartesian product (20 points). In the separate \emph{Burst Management block}, $k=8$ and $\gamma=100$~ms are fixed while $(\theta_u,Q_{\max},W)\in\{0.50,0.90,1.00\}\times\{2,8,32\}\times\{1,3,100\}$ forms a $3\times3\times3$ product (27 points). This block uses 14 base DAG releases, injects a 100\% sustained burst, and keeps offered nodes non-mergeable so that pre-controller merging cannot hide ready-queue backlog from $Q_{\max}$. The default combination is $(k,\gamma)=(8,100\text{ ms})$ and $(\theta_u,Q_{\max},W)=(0.90,8,3)$; the table varies one parameter at a time while holding the others at these values.

\begin{table*}[!t]
\color{black}
\centering
\caption{\rev{Consolidated parameter-sensitivity study. Each row is one concrete factorial run on a one-factor slice through the default combination; all three metrics come from that same run and are truncated to three decimals.}}
\label{tab:parameter_sensitivity}
\scriptsize
\setlength{\tabcolsep}{5pt}
\begin{tabular}{lllccc}
\toprule
Sensitivity group & Parameter & Setting & End-to-end Latency (s) & Deadline Missing Rate & QoE \\
\midrule
\multirow{4}{*}{Moving-average window}
& \multirow{4}{*}{$k$} & 1 & 1.090 & 0.975 & 0.954 \\
& & 4 & 1.093 & 0.977 & 0.951 \\
& & 8 & 1.093 & 0.976 & 0.951 \\
& & 32 & 1.085 & 0.977 & 0.959 \\
\midrule
\multirow{5}{*}{Release-time skew threshold}
& \multirow{5}{*}{$\gamma$} & 0~ms & 1.089 & 0.991 & 0.955 \\
& & 100~ms & 1.093 & 0.976 & 0.951 \\
& & 200~ms & 1.091 & 0.964 & 0.953 \\
& & 500~ms & 1.085 & 0.945 & 0.958 \\
& & 1000~ms & 0.294 & 0.988 & 1.000 \\
\midrule
\multirow{9}{*}{Burst management}
& \multirow{3}{*}{$\theta_u$} & 50\% & 2.211 & 0.436 & 0.985 \\
& & 90\% & 2.212 & 0.484 & 0.983 \\
& & 100\% & 2.218 & 0.436 & 0.977 \\
\cmidrule{2-6}
& \multirow{3}{*}{$Q_{\max}$} & 2 & 2.222 & 0.460 & 0.974 \\
& & 8 & 2.212 & 0.484 & 0.983 \\
& & 32 & 2.274 & 0.579 & 0.924 \\
\cmidrule{2-6}
& \multirow{3}{*}{$W$} & 1 tick & 2.196 & 0.531 & 1.000 \\
& & 3 ticks & 2.212 & 0.484 & 0.983 \\
& & 100 ticks & 2.257 & 0.634 & 0.940 \\
\bottomrule
\end{tabular}
\end{table*}
\FloatBarrier

Along the $\gamma=100$-ms slice, changing the moving-average window is comparatively insensitive: latency spans 1.085--1.093~s, Deadline Missing Rate spans 0.975--0.977, and QoE spans 0.951--0.959. In contrast, the $k=8$ slice shows that $\gamma$ changes which requests are merged and therefore produces clear separation. From 0 to 500~ms, Deadline Missing Rate falls from 0.991 to 0.945 and QoE changes from 0.955 to 0.958, although it is not monotone at every intermediate threshold. At 1000~ms, all five requests become merge-compatible, reducing latency to 0.294~s and yielding QoE 1.000; however, Deadline Missing Rate rises to 0.988. \minorrev{Thus aggressive merging shortens the measured run and its overall lateness without establishing that the offered nodes met their individual deadlines.} Across all 20 concrete refinement runs, QoE spans 0.939--1.000, confirming that the revised workload removes the previous QoE saturation.

The Burst Management block is also non-degenerate: across its 27 concrete points, latency spans 2.188--2.305~s, Deadline Missing Rate spans 0.413--0.698, and QoE spans 0.898--1.000. On the $Q_{\max}=8,W=3$ slice, increasing $\theta_u$ from 50\% to 100\% lowers QoE from 0.985 to 0.977, while Deadline Missing Rate is non-monotone. On the $\theta_u=90\%,W=3$ slice, increasing $Q_{\max}$ from 2 to 32 raises Deadline Missing Rate from 0.460 to 0.579 and lowers QoE from 0.974 to 0.924. With $\theta_u=90\%$ and $Q_{\max}=8$, the 3-tick persistence window trades a lower Deadline Missing Rate than one tick (0.484 versus 0.531) for slightly higher latency and lower QoE; the 100-tick endpoint is worse on all three metrics. Some conservative combinations never activate shedding, while aggressive combinations drop up to 36 of 126 offered nodes. These single-run results establish that the configured parameters are observable under the dedicated workloads; they do not justify treating the defaults as platform-independent optima.
}

\section{Discussion}
\label{sec:discussion}

{\color{black}
\noindent\textbf{Capability boundary.}
Proportional assignment, residual-graph reassignment, EDF/LLF dispatch, and completion-event synchronization apply to an ordinary scheduling DAG. Shared-encoder detection, encoder-output reuse, and decoder-aware merging are MIMONet-specific. A model stage that is not partitioned is scheduled as one non-partitionable task. This paper does not implement Transformer-specific partitioning, KV-cache-aware scheduling, or intra-kernel preemption. Although the generic components are architecturally applicable beyond MIMONet, we do not claim isolated empirical generalization to ordinary independent-DNN DAGs.

\noindent\textbf{Guarantee boundary.}
The paper intentionally distinguishes three regimes. The admitted refined-DAG regime has a conditional composition result. For a non-partitionable task, Eq.~\eqref{eq:nonpart_deadline} defines a profile-bounded admission threshold; empirical timing evidence additionally requires direct per-task verification within NP1--NP3, which the aggregate ratio table does not supply. Overload shedding is best-effort priority-aware degradation and may miss deadlines. \minorrev{QoE quantifies run-level termination lateness, while DMR counts unsuccessful offered-node outcomes; neither high QoE nor shorter termination time compensates for dropped or skipped service.}

\noindent\textbf{Scalability.}
Height construction and proportional assignment are $O(|V|+|E|)$ per DAG. A direct pairwise merge search over a frontier $S$ is $O(|S|^2)$, or $O(|S|\log|S|)$ after sorting by release time and sweeping a fixed skew window. These CPU-side costs do not establish accelerator capacity; larger branch counts, more co-running DAGs, and non-partitionable tasks can still invalidate admission. Table~\ref{tab:combined_overhead} reports the measured runtime and memory overhead on four embedded platforms.

\noindent\textbf{Limitations and future work.}
\minorrev{The current implementation schedules workloads on one accelerator using calibrated execution estimates and benefits most from MIMONet-like sharing. It has no general per-level admission certificate or separately validated contention bound; the conditional non-partitionable rule is therefore not established by the aggregate ratio study.} No guarantee is claimed once sustained burst demand violates admission. Every reported experiment campaign is validated before insertion. Empirical evaluation on ordinary independent-DNN DAGs, Transformer-specific decomposition, KV-cache-aware scheduling, multi-accelerator placement, energy/thermal control, direct non-partitionable bound validation, and formal burst-overload guarantees remain future work.}

\section{Related Work}

\noindent\textbf{Real-time DAG scheduling.} DAG workloads have long been studied in both server platforms~\citep{ahmed2022exact,sakellariou2004hybrid,wu2015hierarchical,canon2008comparative,panahi2009framework} and embedded settings~\citep{xie2015heterogeneity,bi2022response,zhang2024boxr}, with foundational results traceable to Graham’s classic bound~\citep{graham1969bounds}. Recent work advances schedulability analysis and runtime support for parallel, real-time DAGs~\citep{ueter2018reservation,baruah2015federated,jiang2020real,jiang2017semi,li2014analysis,li2013outstanding,melani2015response,bi2022response,he2022bounding,chen2023precise}. Outside real-time robotics, systems for {evolving graphs} (e.g., CommonGraph~\citep{10.1145/3575693.3575713,10.1145/3597635.3598022} and the MEGA accelerator~\citep{10.1145/3613424.3614260}) exploit temporal locality, work sharing, and redundancy removal as the graph changes. While their goals are high-throughput analytics rather than deadline-aware execution, the underlying idea, i.e., reuse under dynamic topology, echoes our MIMONet-aware refinement that amortizes shared encoder computation in changing DAGs. Yet, most existing schedulers still assume static graphs or lack support for the multi-input multi-output structure and dynamically evolving workloads characteristic of MIMONet, which limits their applicability to our problem. We therefore include a representative technique, intermediate deadline assignment~\citep{panahi2009framework, liu2000real,li2023rt}, as a baseline capable of coping with certain forms of dynamicity.

{\color{black}
To strengthen the comparison beyond EDF and internal ablations, the evaluation adapts least-laxity-first scheduling~\citep{liu2000real} to a mutating DAG by subtracting the profiled remaining critical-path cost from each eligible node's absolute deadline. LLF-DAG receives the same current graph and profiles as \tool, does not see future arrivals, and does not use MIMONet-specific merge/reuse logic. Unit tests validate the remaining-critical-path calculation and runtime priority order, and LLF-DAG is included in the validated overall, non-partitionable-task, and burst-task comparisons.}

{\color{black}
\noindent\textbf{Mixed-criticality scheduling.} The mixed-criticality scheduling (MCS) literature, surveyed by Burns and Davis~\citep{burns2017mixedcrit}, assigns design-time criticality classes and provides mode-switch analyses. \minorrev{\textsc{MERIS} instead uses the release-relative deadline-urgency score in Eq.~\eqref{eq:crit_score} within a configured threshold/protection policy. The score contains no class-criticality input; explicit droppability, thresholds, and any class ordering are separate policy inputs (Sec.~\ref{sec:burst_task_scenarios}).} It has no \textsc{LO}$\to$\textsc{HI} mode-switch proof and is not presented as an MCS guarantee; the dropped-task composition disclosed in Sec.~\ref{sec:eval_burstmanagement} is a diagnostic of best-effort degradation.}

\noindent\textbf{Real-time scheduling in ROS.} A series of ROS-focused frameworks has explored dynamic priorities, end-to-end latency control, multicore-aware policies, and feedback-based adaptation~\citep{DBLP:conf/rtss/JiangJGLTW22,DBLP:conf/rtss/LiGJGDL22,DBLP:conf/rtss/TeperGUBC22,DBLP:conf/rtss/BlassCBB21,DBLP:conf/rtss/TangFG0LD020,DBLP:conf/rtas/ChoiXK21,DBLP:conf/rtas/BlassHLZB21,sobhani2023timing,li2025mace}. While effective for conventional models, these systems generally do not natively support the dynamic, MIMO-based modern DNN pipelines. Our design addresses this gap by creating a scheduler tailored for MIMONet workloads within ROS.

{\color{black}
\noindent\textbf{Real-time DNN inference.} Prior work achieves low-latency inference via approximation~\citep{zhou2018s}, streaming/scheduling frameworks~\citep{predjoule,LI2021101936}, compiler optimization~\citep{chen2025your,chen2023dycl}, and OS/runtime support~\citep{kang2021lalarand,bib:RTSS19:Xiang}. These systems motivate stronger scheduling baselines but do not provide the same explicit shared-encoder/decoder mapping used here. Our contribution is scoped to that MIMONet-aware mapping plus the conditional and empirical regimes stated in Sec.~\ref{sec:schedulability}; it is not a claim of dominance over every static-DNN runtime.}

\noindent\textbf{Weight sharing and parameter efficiency.} Weight sharing reduces compute and memory in both single-task~\citep{bib:ICLR19:Yu,bib:ICML17:Bolukbas} and multi-task~\citep{bib:NIPS18:He,bib:TMC21:He,bib:MobiSys20:Lee} settings. Recent graph-rewriting strategies further “stitch’’ multiple weight-shared networks into unified graphs to enable efficient multi-task inference~\citep{wang2022stitching}. Complementary to weight sharing, parameter-efficient fine-tuning (PEFT) limits the number of trainable parameters during model adaptation (surveyed in~\citep{han2024parameter}) and can be realized in distributed settings via LoRA-style adapters (e.g., DLora~\citep{gao2024dlora}). These PEFT techniques primarily target training-time adaptation and deployment scalability; our contribution is orthogonal, focusing on inference-time scheduling and deadline satisfaction for weight-shared MIMONet pipelines on embedded robots.

\noindent\textbf{Energy-, power-, and thermal-aware systems.} Although our focus is on inference-time real-time scheduling for embedded robots, there is complementary work on energy/thermal efficiency at infrastructure and training layers. Warm-water cooling and heat-recycling techniques improve datacenter and edge-datacenter efficiency and hotspot management~\citep{jiang2019fine,pei2022cooledge,zhu2020heat,chen2022nmtsloth, chen2022nicgslowdown, feng2024llmeffichecker, zhang2024mii}. Separately, power-aware collective communication reduces energy cost during LLM training~\citep{jia2024pccl}. These directions optimize energy and thermals rather than end-to-end deadlines, but the underlying theme, aligning computation with physical constraints, resonates with our design goal of respecting resource limits while sustaining performance in embedded deployments.

\minorrev{\noindent\textbf{Adaptive on-device learning and serving.} Orion adapts memory allocation across online continual-learning components~\citep{li2026orion}. TetraRL jointly considers timing, reward, memory, and energy for on-device reinforcement learning~\citep{li2026tetrarl}. These systems allocate resources for adaptation, whereas \tool schedules changing inference DAGs and reuses shared encoder computation. Tri-serve combines dependency-, arithmetic-intensity-, and thermal-aware DVFS for energy-efficient multimodal serving~\citep{jia2026triserve}, complementing our deadline- and precedence-aware dispatch.}

\noindent\textbf{Intelligent robotic systems.} Lightweight, real-time deep learning has been explored for both on-robot training~\citep{yoo2022learning,shirvani2024duojoule,li2023mathrm,li2025lemix} and inference~\citep{popov2022nvradarnet,li2024genie,li2023rt,li2023red,li2025lemix,shirvani2024duojoule,li2023mathrm}, frequently in safety-critical settings with explicit timing requirements, including multi-robot coordination~\citep{nikkhoo2023pimbot,he2023robust}, UAV planning~\citep{xu2022dpmpc}, geo-localization~\citep{zhang2022learning}, tracking~\citep{botros2022fully}, HCI~\citep{lee2022towards}, and autonomous driving~\citep{he2022robust,he2023robustev}. Scheduling frameworks have also incorporated confidence-awareness for complex tasks such as multi-object tracking~\citep{kang2022rt}. Our work fits this trajectory but specifically targets real-time execution of MIMO DNNs in robotic systems.

\section{Conclusion}
{\color{black}
We presented \tool as a MIMONet-aware framework for model scheduling under real-time constraints that maps an application DAG and its internal encoder/decoder graphs to one refined scheduling DAG. Proportional intermediate deadlines compose to the end-to-end deadline only when the stated execution, release, precedence, and level-service premises hold. \minorrev{Non-partitionable tasks remain indivisible scheduling nodes with configured budgets. The profile-bounded admission rule under NP1--NP3 is a conditional deployment requirement, and the ratio study supplies comparative observations rather than verification of that rule.} Overload shedding is best-effort outside the admitted regime. The QoE implementation uses consistent time units and honors the configured $\lambda$. \minorrev{QoE measures the continuous lateness of run termination, while DMR retains skipped, dropped, late, and erroneous outcomes in the offered-node population. Neither metric alone establishes task completion quality or safety.} The overall-effectiveness studies apply a common workload to all compared methods on Nano Orin, AGX Orin, and M3 Max; the non-partitionable-task, burst-task, and parameter-sensitivity studies report additional validated results on these platforms.}

\appendix
\section{\minorrev{Descriptive Variability of Existing Overall-Effectiveness Runs}}
\label{sec:existing_variability}
\minorrev{Table~\ref{tab:existing_variability} summarizes the same August 2026 records used to regenerate Tables~\ref{tab:extended_overall_ablation} and~\ref{tab:extended_overall_effectiveness}. Each group contains seeds 101, 202, 303, 404, and 505 under one frozen configuration. All 225 original manifests report complete execution with matching result hashes; no run was added or excluded in this reanalysis. The table includes the external baselines and the complete system. Main-table DMR/QoE values remain those of the median-latency run, while the ranges here use all five observations. Zero-miss counts use raw task outcomes, not rounded or truncated DMR values. The single-run burst and sensitivity conditions elsewhere in the paper do not acquire repetition-based uncertainty estimates from this table.}
\input{tables/overall_variability.tex}
\FloatBarrier

\reinbibliography{references}
\end{document}

%% file: paper-preamble.tex
\def\MinorCleanVersion{1}

\newcommand{\rev}[1]{{\color{black}#1}}
\definecolor{revisionred}{RGB}{200,0,0}
\ifdefined\MinorCleanVersion
  \colorlet{revisioncolor}{black}
\else
  \colorlet{revisioncolor}{revisionred}
\fi
\newcommand{\minorrev}[1]{{\color{revisioncolor}#1}}

\usepackage{tikz}
\newcommand{\takeaway}[2]{
\par\noindent
\begin{tikzpicture}
\node[
 draw=black, line width=0.9pt, fill=white,
 rounded corners=1.2pt,
 inner xsep=6pt, inner ysep=6pt,
 text width=\dimexpr0.97\columnwidth-12pt\relax,
 align=left, font=\small
] (tkbody) {\rule[-0.3ex]{0pt}{3.4ex}\ignorespaces #2};
\node[
 fill=white, text=black, font=\footnotesize\bfseries,
 inner xsep=6pt, inner ysep=2.5pt,
 anchor=south west
] (tkhead) at ([xshift=4pt,yshift=0.3pt]tkbody.north west) {\strut #1};
\end{tikzpicture}\par
}
\def \Approach{\rev{\textsc{MERIS}}\xspace}
\def \tool{\rev{\textsc{MERIS}}\xspace}

\usepackage{booktabs}
\usepackage{amsfonts}

\usepackage{graphicx}
\usepackage{hyperref}
\usepackage{textcomp}
\usepackage{listings}
\usepackage{adjustbox}
\usepackage{xspace}    
\usepackage{multirow}
\usepackage{multicol}
\usepackage{xcolor} 
\usepackage{tcolorbox}
\usepackage{wrapfig}
\usepackage{diagbox}
\usepackage{amsmath}
\usepackage{pifont}
\usepackage{breakcites}
\usepackage{array}
\usepackage{framed}
\definecolor{shadecolor}{rgb}{0.92,0.92,0.92}

\usepackage[algo2e,linesnumbered,boxed,ruled]{algorithm2e}
\usepackage{algpseudocode}
\usepackage{listings}
\usepackage{bbding}

\usepackage{enumitem}
\usepackage{subcaption}

\usepackage{placeins}
\providecommand{\Description}[1]{}

\colorlet{REINPrimary}{black}
\colorlet{REINInk}{black}
\colorlet{REINForest}{black}

%% file: paper-info.tex
\title{MERIS: Adaptive Real-Time Scheduling of Refined Inference DAGs in Dynamic Robotic Environments}
\author{Zexin Li\textsuperscript{1,*}\reinorcid{0000-0001-8758-2151}\quad
Tao Ren\textsuperscript{2,*}\reinorcid{0009-0007-9638-8386}\quad Johnathan Liu\textsuperscript{3}\\
Xiaoxi He\textsuperscript{4}\reinorcid{0009-0001-1178-263X}\quad Cong Liu\textsuperscript{5}\reinorcid{0009-0001-6775-1269}}
\reinaffiliation{\textsuperscript{1}Nanyang Technological University, Singapore\quad
\textsuperscript{2}Rein Protocol Foundation\\[2pt]
\textsuperscript{3}University of Maryland, Baltimore County\quad
\textsuperscript{4}University of Macau\\[2pt]
\textsuperscript{5}University of California, Riverside}
\reinemails{zexin.li@ntu.edu.sg; tempest.ren@rein-protocol.org; jliu13@umbc.edu;\\hexiaoxi@um.edu.mo; congl@ucr.edu}
\reinshorttitle{MERIS: Adaptive Real-Time Scheduling}
\reintrack{REAL-TIME SYSTEMS / ROBOTIC INFERENCE}
\reinstatus{RESEARCH PREPRINT}
\reinkeywords{Autonomous Embedded Systems; Deep Neural Networks}
\makeatletter
\apptocmd{\rein@brandrow}{%
  \noindent
  \begin{minipage}[c]{.23\textwidth}\centering
  \includegraphics[width=\linewidth,height=30pt,keepaspectratio]{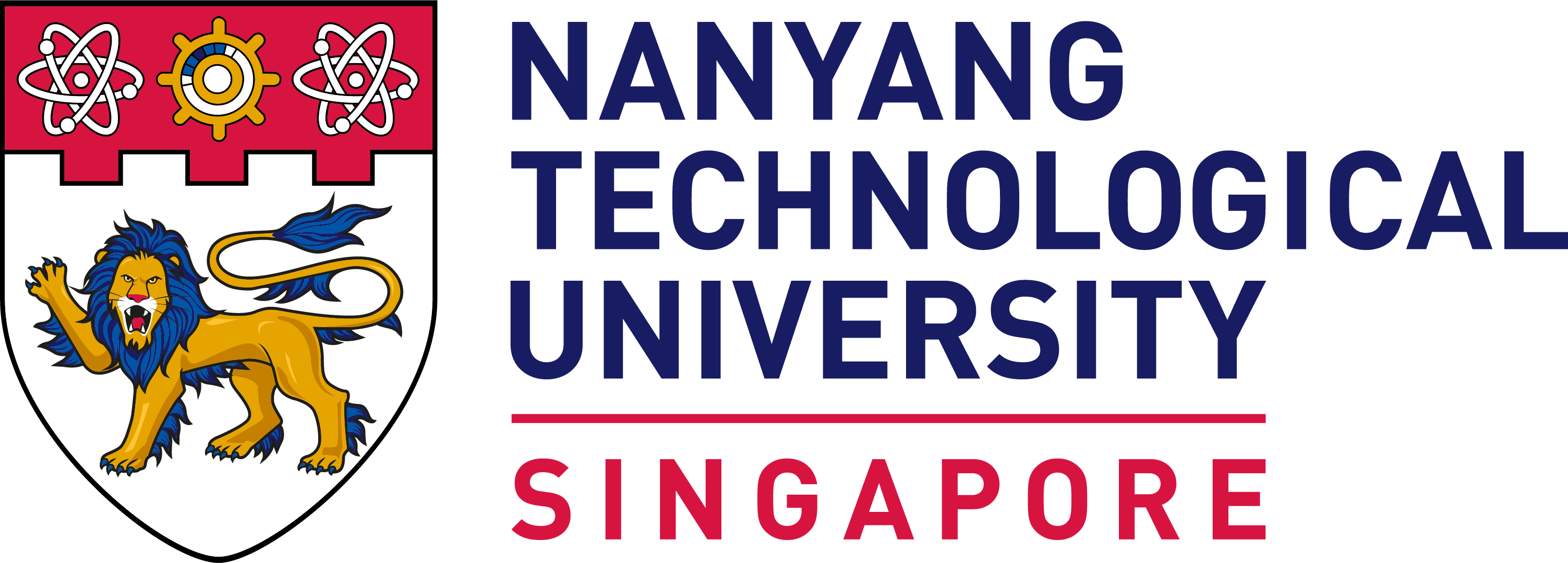}\end{minipage}\hfill
  \begin{minipage}[c]{.23\textwidth}\centering
  \includegraphics[width=\linewidth,height=30pt,keepaspectratio]{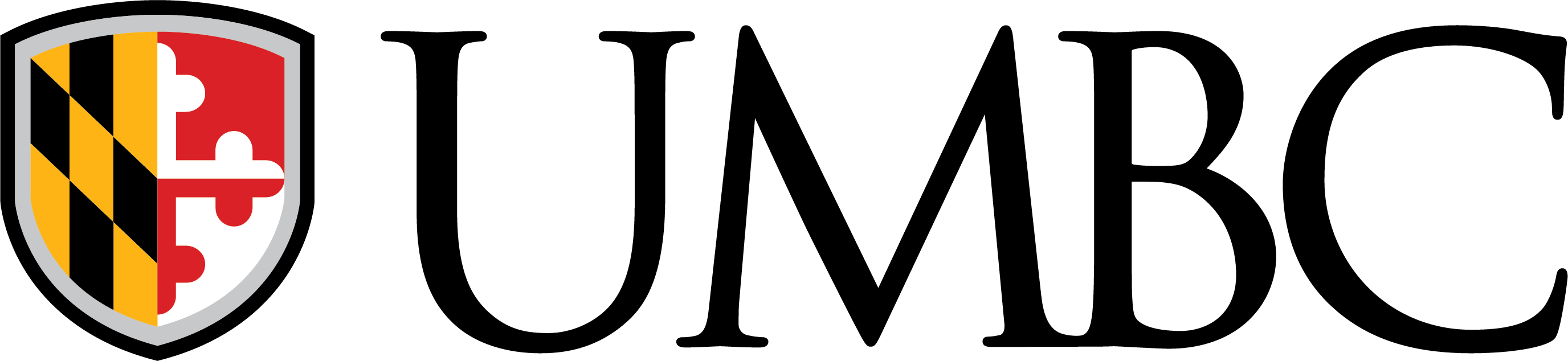}\end{minipage}\hfill
  \begin{minipage}[c]{.23\textwidth}\centering
  \includegraphics[width=\linewidth,height=30pt,keepaspectratio,trim=30bp 145bp 20bp 145bp,clip]{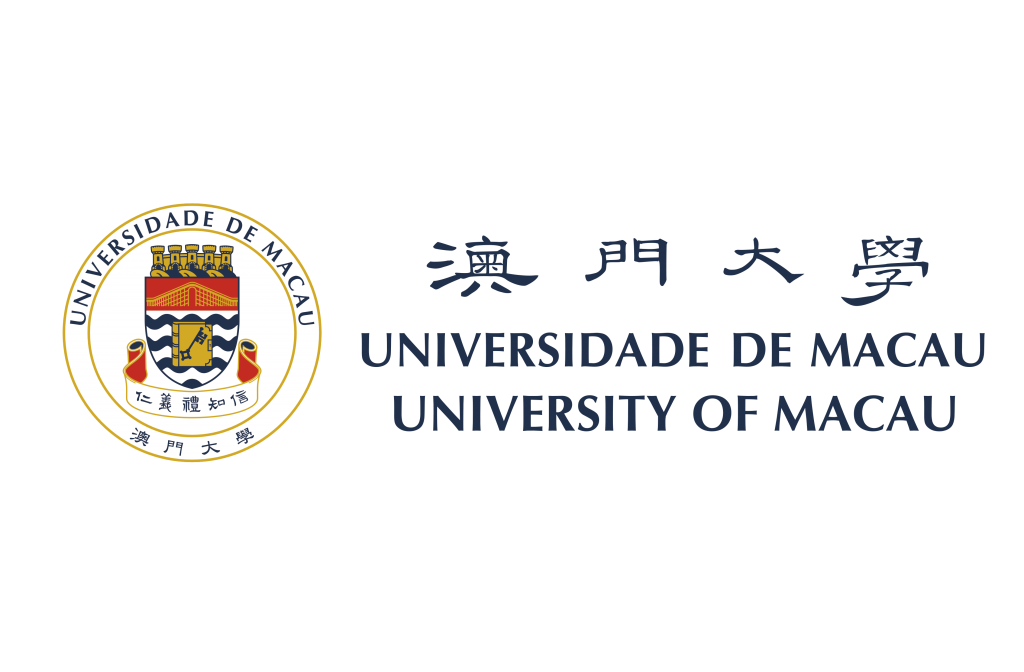}\end{minipage}\hfill
  \begin{minipage}[c]{.23\textwidth}\centering
  \includegraphics[width=\linewidth,height=30pt,keepaspectratio]{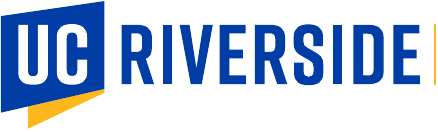}\end{minipage}\par
  \kern 7pt 
}{}{\PackageError{meris}{Could not extend the institution logo row}{}}
\makeatother

\apptocmd{\maketitle}{%
  \begingroup
  \footnotetext[1]{These authors contributed equally to this work.}%
  \endgroup
}{}{\PackageError{meris}{Could not add the equal-contribution footnote}{}}

%% file: tables/overall_tables.tex
\begin{table*}[!t]
\color{black}
\centering
\caption{\rev{Ablation study on the extended minibenchmarks under platform-appropriate Tight TaskConfigs. EDF is the base scheduler; \tool-FG adds MIMONet refinement, \tool-IDA adds residual-graph intermediate-deadline reassignment, and \tool adds event-driven synchronization and overload control. End-to-end latency, Deadline Missing Rate, and QoE are reported in separate blocks. \minorrev{Each entry is the complete tuple from the median-latency run among five existing repetitions; red cells correct the previous run selection or formatting.}}}
\label{tab:extended_overall_ablation}
\small
\resizebox{\textwidth}{!}{%
\begin{tabular}{l|cccc|cccc|cccc}
\hline
\multirow{2}{*}{Platform}
& \multicolumn{4}{c|}{Urban Navigation}
& \multicolumn{4}{c|}{Emergency Obstacle Avoidance}
& \multicolumn{4}{c}{Night Driving} \\
\cline{2-13}
& EDF & \tool-FG & \tool-IDA & \tool
& EDF & \tool-FG & \tool-IDA & \tool
& EDF & \tool-FG & \tool-IDA & \tool \\
\hline
\multicolumn{13}{c}{End-to-end Latency (s)} \\
\hline
Nano Orin
& 2.418 & 2.269 & 2.410 & \textbf{1.624}
& 1.872 & 1.744 & 1.859 & \textbf{1.495}
& \minorrev{\textbf{1.196}} & 1.198 & 1.210 & \minorrev{1.202} \\
AGX Orin
& 7.011 & 7.012 & 9.449 & \textbf{2.728}
& 3.787 & 2.850 & \minorrev{4.389} & \minorrev{\textbf{2.763}}
& 3.168 & \minorrev{\textbf{2.179}} & \minorrev{3.067} & \minorrev{2.392} \\
M3 Max
& 13.266 & 10.797 & 13.430 & \textbf{10.790}
& 10.256 & 9.049 & 10.298 & \textbf{9.046}
& 14.590 & 11.366 & 14.596 & \textbf{7.164} \\
\hline
\multicolumn{13}{c}{Deadline Missing Rate} \\
\hline
Nano Orin
& 0.600 & \textbf{0.000} & 0.480 & \textbf{0.000}
& 0.650 & \textbf{0.000} & 0.575 & \textbf{0.000}
& \minorrev{\textbf{0.000}} & 0.200 & 0.500 & \minorrev{0.600} \\
AGX Orin
& 0.020 & \textbf{0.000} & 0.960 & \textbf{0.000}
& 0.925 & \minorrev{\textbf{0.525}} & \minorrev{0.975} & \minorrev{0.775}
& 0.966 & \minorrev{\textbf{0.733}} & \minorrev{0.933} & \minorrev{0.966} \\
M3 Max
& 0.832 & \textbf{0.000} & 0.841 & \textbf{0.000}
& 0.631 & \textbf{0.000} & 0.607 & \textbf{0.000}
& 0.890 & \textbf{0.000} & 0.868 & \textbf{0.000} \\
\hline
\multicolumn{13}{c}{QoE} \\
\hline
Nano Orin
& 0.918 & \textbf{1.000} & 0.926 & \textbf{1.000}
& 0.916 & \textbf{1.000} & 0.928 & \textbf{1.000}
& \minorrev{\textbf{0.997}} & 0.995 & 0.983 & \minorrev{0.991} \\
AGX Orin
& \textbf{1.000} & \textbf{1.000} & 0.302 & \textbf{1.000}
& 0.534 & \textbf{1.000} & \minorrev{0.399} & \textbf{1.000}
& 0.514 & \textbf{1.000} & \minorrev{0.544} & \minorrev{0.887} \\
M3 Max
& 0.338 & \textbf{1.000} & 0.319 & \textbf{1.000}
& 0.609 & \textbf{1.000} & 0.593 & \textbf{1.000}
& 0.259 & \textbf{1.000} & 0.258 & \textbf{1.000} \\
\hline
\end{tabular}
}
\end{table*}

\begin{table*}[!t]
\color{black}
\centering
\caption{\rev{External-baseline comparison on the extended minibenchmarks under platform-appropriate Tight TaskConfigs. EDF and LLF-DAG are compared with the complete \tool system. End-to-end latency, Deadline Missing Rate, and QoE are reported in separate blocks. \minorrev{Each entry is the complete tuple from the median-latency run among five existing repetitions; red cells correct the previous run selection or formatting.}}}
\label{tab:extended_overall_effectiveness}
\small
\resizebox{\textwidth}{!}{%
\begin{tabular}{l|ccc|ccc|ccc}
\hline
\multirow{2}{*}{Platform}
& \multicolumn{3}{c|}{Urban Navigation}
& \multicolumn{3}{c|}{Emergency Obstacle Avoidance}
& \multicolumn{3}{c}{Night Driving} \\
\cline{2-10}
& EDF & LLF-DAG & \tool
& EDF & LLF-DAG & \tool
& EDF & LLF-DAG & \tool \\
\hline
\multicolumn{10}{c}{End-to-end Latency (s)} \\
\hline
Nano Orin
& 2.418 & 2.405 & \textbf{1.624}
& 1.872 & 1.872 & \textbf{1.495}
& \minorrev{1.196} & \minorrev{\textbf{1.195}} & \minorrev{1.202} \\
AGX Orin
& 7.011 & 8.339 & \textbf{2.728}
& 3.787 & \minorrev{13.842} & \minorrev{\textbf{2.763}}
& 3.168 & 2.671 & \minorrev{\textbf{2.392}} \\
M3 Max
& 13.266 & 13.450 & \textbf{10.790}
& 10.256 & 10.394 & \textbf{9.046}
& 14.590 & 14.648 & \textbf{7.164} \\
\hline
\multicolumn{10}{c}{Deadline Missing Rate} \\
\hline
Nano Orin
& 0.600 & 0.480 & \textbf{0.000}
& 0.650 & 0.700 & \textbf{0.000}
& \minorrev{\textbf{0.000}} & \minorrev{\textbf{0.000}} & \minorrev{0.600} \\
AGX Orin
& 0.020 & 0.920 & \textbf{0.000}
& 0.925 & \minorrev{1.000} & \minorrev{\textbf{0.775}}
& 0.966 & \minorrev{\textbf{0.866}} & \minorrev{0.966} \\
M3 Max
& 0.832 & 0.821 & \textbf{0.000}
& 0.631 & 0.680 & \textbf{0.000}
& 0.890 & 0.881 & \textbf{0.000} \\
\hline
\multicolumn{10}{c}{QoE} \\
\hline
Nano Orin
& 0.918 & 0.930 & \textbf{1.000}
& 0.916 & 0.917 & \textbf{1.000}
& \minorrev{0.997} & \minorrev{\textbf{0.997}} & \minorrev{0.991} \\
AGX Orin
& \textbf{1.000} & 0.467 & \textbf{1.000}
& 0.534 & \minorrev{0.080} & \textbf{1.000}
& 0.514 & 0.704 & \minorrev{\textbf{0.887}} \\
M3 Max
& 0.338 & 0.317 & \textbf{1.000}
& 0.609 & 0.560 & \textbf{1.000}
& 0.259 & 0.255 & \textbf{1.000} \\
\hline
\end{tabular}
}
\end{table*}

%% file: tables/overall_variability.tex
\begin{table*}[!htbp]
\color{revisioncolor}
\centering\small
\caption{\minorrev{Variability across the five existing repetitions per platform--scenario--method group. Latency is shown as median [minimum, maximum]; the other columns are ranges across all five repetitions, not uncertainty attached to the selected tuple. Values are truncated to three decimals.}}
\label{tab:existing_variability}
\resizebox{\textwidth}{!}{%
\begin{tabular}{lllrllr}
\toprule
Platform & Scenario & Method & Latency median [min, max] (s) & DMR [min, max] & Run-level QoE [min, max] & Zero-miss runs \\
\midrule
Nano Orin & Urban & EDF & 2.418 [2.410, 2.466] & [0.580, 0.680] & [0.878, 0.926] & 0/5 \\
 &  & LLF-DAG & 2.405 [2.400, 2.412] & [0.420, 0.500] & [0.924, 0.935] & 0/5 \\
 &  & \tool & 1.624 [1.602, 1.636] & [0.000, 0.000] & [1.000, 1.000] & 5/5 \\
\addlinespace[2pt]
Nano Orin & Emergency & EDF & 1.872 [1.859, 1.889] & [0.625, 0.700] & [0.901, 0.927] & 0/5 \\
 &  & LLF-DAG & 1.872 [1.855, 1.895] & [0.500, 0.700] & [0.897, 0.931] & 0/5 \\
 &  & \tool & 1.495 [1.468, 1.503] & [0.000, 0.000] & [1.000, 1.000] & 5/5 \\
\addlinespace[2pt]
Nano Orin & Night & EDF & 1.196 [1.187, 1.208] & [0.000, 0.600] & [0.984, 1.000] & 3/5 \\
 &  & LLF-DAG & 1.195 [1.185, 1.224] & [0.000, 0.733] & [0.969, 1.000] & 3/5 \\
 &  & \tool & 1.202 [1.184, 1.227] & [0.000, 0.866] & [0.965, 1.000] & 1/5 \\
\addlinespace[2pt]
\midrule
AGX Orin & Urban & EDF & 7.011 [7.011, 15.993] & [0.020, 0.940] & [0.098, 1.000] & 0/5 \\
 &  & LLF-DAG & 8.339 [7.207, 16.438] & [0.900, 0.980] & [0.093, 1.000] & 0/5 \\
 &  & \tool & 2.728 [1.852, 5.502] & [0.000, 0.100] & [1.000, 1.000] & 3/5 \\
\addlinespace[2pt]
AGX Orin & Emergency & EDF & 3.787 [3.512, 5.913] & [0.750, 0.950] & [0.243, 0.632] & 0/5 \\
 &  & LLF-DAG & 13.842 [5.172, 15.414] & [0.875, 1.000] & [0.070, 0.300] & 0/5 \\
 &  & \tool & 2.763 [2.067, 4.067] & [0.325, 0.925] & [0.461, 1.000] & 0/5 \\
\addlinespace[2pt]
AGX Orin & Night & EDF & 3.168 [2.175, 5.982] & [0.000, 1.000] & [0.204, 1.000] & 1/5 \\
 &  & LLF-DAG & 2.671 [2.295, 3.283] & [0.866, 0.966] & [0.484, 0.975] & 0/5 \\
 &  & \tool & 2.392 [1.595, 3.610] & [0.433, 0.966] & [0.415, 1.000] & 0/5 \\
\addlinespace[2pt]
\midrule
M3 Max & Urban & EDF & 13.266 [13.196, 13.415] & [0.796, 0.832] & [0.321, 0.347] & 0/5 \\
 &  & LLF-DAG & 13.450 [13.352, 13.524] & [0.821, 0.846] & [0.310, 0.328] & 0/5 \\
 &  & \tool & 10.790 [10.789, 10.791] & [0.000, 0.000] & [1.000, 1.000] & 5/5 \\
\addlinespace[2pt]
M3 Max & Emergency & EDF & 10.256 [10.159, 10.397] & [0.593, 0.662] & [0.559, 0.650] & 0/5 \\
 &  & LLF-DAG & 10.394 [10.325, 10.469] & [0.647, 0.698] & [0.536, 0.583] & 0/5 \\
 &  & \tool & 9.046 [9.046, 9.047] & [0.000, 0.000] & [1.000, 1.000] & 5/5 \\
\addlinespace[2pt]
M3 Max & Night & EDF & 14.590 [14.472, 14.630] & [0.852, 0.890] & [0.256, 0.267] & 0/5 \\
 &  & LLF-DAG & 14.648 [14.599, 14.878] & [0.870, 0.888] & [0.240, 0.258] & 0/5 \\
 &  & \tool & 7.164 [7.092, 7.223] & [0.000, 0.000] & [1.000, 1.000] & 5/5 \\
\addlinespace[2pt]
\bottomrule
\end{tabular}}
\end{table*}